\documentclass[10pt,twocolumn,twoside, romanappendices]{IEEEtran}
\usepackage{graphicx}
\usepackage{subfigure}
\usepackage{amsmath}
\usepackage{amssymb}
\usepackage{tabulary}
\usepackage{multirow}
\usepackage{color}
\usepackage{algpseudocode}
\usepackage{algorithm,float}
\usepackage{xcolor}
\usepackage{blindtext}
\usepackage{cite}
\usepackage{booktabs}
\usepackage{color}
\usepackage{array}
\usepackage{tabularx}
\usepackage{algpseudocode}
\usepackage{comment}
\usepackage{pgfplots}
\usepackage{tikz}
\usepackage{caption}    

\definecolor{bblue}{HTML}{4F81BD}
\definecolor{rred}{HTML}{C0504D}
\definecolor{ggreen}{HTML}{9BBB59}
\definecolor{ppurple}{HTML}{9F4C7C}

\definecolor{dbcolor}{rgb}{0,0,1}

\begin{document}
\title{Fast 2-D Complex Gabor Filter with Kernel Decomposition}

\author{Suhyuk Um$^\dag$, Jaeyoon Kim$^\dag$, and Dongbo Min, \IEEEmembership{Senior Member, IEEE}

\thanks{S. Um, J. Kim and D. Min are with the Department of Computer Science and Engineering in Chungnam National University,
Korea. (e-mail: suhyuk1104@gmail.com; wodbs135@naver.com;
dbmin@cnu.ac.kr). $^\dag$: Two authors contribute equally to this
work.}}

\markboth{IEEE Transaction on Image Processing}%
{\MakeLowercase{}}
%

%
%
%

\IEEEcompsoctitleabstractindextext{%
\begin{abstract}
2-D complex Gabor filtering has found numerous applications in the
fields of computer vision and image processing. Especially, in some
applications, it is often needed to compute 2-D complex Gabor filter
bank consisting of the 2-D complex Gabor filtering outputs at
multiple orientations and frequencies. Although several approaches
for fast 2-D complex Gabor filtering have been proposed, they
primarily focus on reducing the runtime of performing the 2-D
complex Gabor filtering once at specific orientation and frequency.
To obtain the 2-D complex Gabor filter bank output, existing methods
are repeatedly applied with respect to multiple orientations and
frequencies. In this paper, we propose a novel approach that
efficiently computes the 2-D complex Gabor filter bank by reducing
the computational redundancy that arises when performing the Gabor
filtering at multiple orientations and frequencies. The proposed
method first decomposes the Gabor basis kernels to allow a fast
convolution with the Gaussian kernel in a separable manner. This
enables reducing the runtime of the 2-D complex Gabor filter bank by
reusing intermediate results of the 2-D complex Gabor filtering
computed at a specific orientation. Furthermore, we extend this idea
into 2-D localized sliding discrete Fourier transform (SDFT) using
the Gaussian kernel in the DFT computation, which lends a spatial
localization ability as in the 2-D complex Gabor filter.
Experimental results demonstrate that our method runs faster than
state-of-the-arts methods for fast 2-D complex Gabor filtering,
while maintaining similar filtering quality.
\end{abstract}

}

\maketitle
\IEEEdisplaynotcompsoctitleabstractindextext
\IEEEpeerreviewmaketitle

\section{Introduction}
2-D complex Gabor filter has been widely used in numerous
applications of computer vision and image processing thanks to its
elegant properties of extracting locally-varying structures from an
image. In general, it is composed of the Gaussian kernel and complex
sinusoidal modulation term, which can be seen as a special case of
short time discrete Fourier transform (STFT). The Gabor basis
functions defined for each pixel offer good spatial-frequency
localization 
\cite{tit/Daubechies90}.

It was first discovered in \cite{josa85} that the 2D receptive field
profiles of simple cells in the mammalian visual cortex can be
modeled by a family of Gabor filters. It was also known in
\cite{ivc/ShenBF07,VQML-15} that image analysis approaches based on
the Gabor filter conceptually imitate the human visual system (HVS).
The 2-D complex Gabor filter is invariant to rotation, scale,
translation and illumination \cite{tip/KamarainenKK06}, and it is
particularly useful for extracting features at a set of
different orientations and frequencies from the image. 
Thanks to such properties, it has found a great variety of
applications in the field of computer vision and image processing,
including texture analysis \cite{pr/WeldonHD96, TexturePR07, TextureTIP09, tip/LiDZ15}, 
face recognition \cite{FR-Gabor97, FR-NN03, FR-PAA06, FR-FG08, FR-IJMLC16}, 
face expression recognition \cite{pr/GuXVHL12,Facial-15} and
fingerprint recognition \cite{Kasban2016910}.

Performing the 2-D complex Gabor filtering for all pixels over an
entire image, however, often provokes a heavy computational cost.
With the 2-D complex Gabor kernel defined at specific orientation
and frequency, the filtering is performed by moving a reference
pixel to be filtered one pixel at a time. The complex kernel hinders
the fast computation of the 2-D complex Gabor filtering in the
context similar to edge-aware filters \cite{GuidedFilter-ECCV10,
DomainTrans-TOG11, FGS-TIP14} that are widely used in numerous
computer vision applications.

To expedite the 2-D complex Gabor filtering, several efforts have
been made, for instance, by making use of the fast Fourier transform
(FFT), infinite impulse response (IIR) filters, or finite impulse
response (FIR) filters \cite{sigpro/QiuZC99, jei/NestaresNPT98,
tsp/YoungVG02, tip/BernardinoS06}. It is shown in
\cite{sigpro/QiuZC99} that the Gabor filtering and synthesis for a
1-D signal consisting of $N$ samples can be performed with the same
complexity as the FFT, $O(NlogN)$. In \cite{jei/NestaresNPT98},
separable FIR filters are applied to implement fast 2-D complex
Gabor filtering by exploiting particular relationships between the
parameters of the 2-D complex Gabor filter in a multiresolution
pyramid. The fast pyramid implementation is, however, feasible only
for the particular setting of Gabor parameters, e.g., scale of $2^i$
with an integer $i$. Young \emph{et al.} \cite{tsp/YoungVG02}
proposes to formulate the 2-D complex Gabor filter as IIR filters
that efficiently work in a recursive manner. They decompose the
Gabor filter with multiple IIR filters through z-transform, and then
performs the recursive filtering in a manner similar to recursive Gaussian filtering \cite{icpr/VlietYV98}. 
To the best of our knowledge, the fastest algorithm for the 2-D
complex Gabor filtering is the work of Bernardino and Santos-Victor
\cite{tip/BernardinoS06} that decomposes the 2-D complex Gabor
filtering into more efficient Gaussian filtering and sinusoidal
modulations. It was reported in \cite{tip/BernardinoS06} that this
method reduces up to $39\%$ the number of arithmetic operations
compared to the recursive Gabor filtering \cite{tsp/YoungVG02}.

\begin{figure}[t]
\centering
\includegraphics[width=0.48\textwidth]{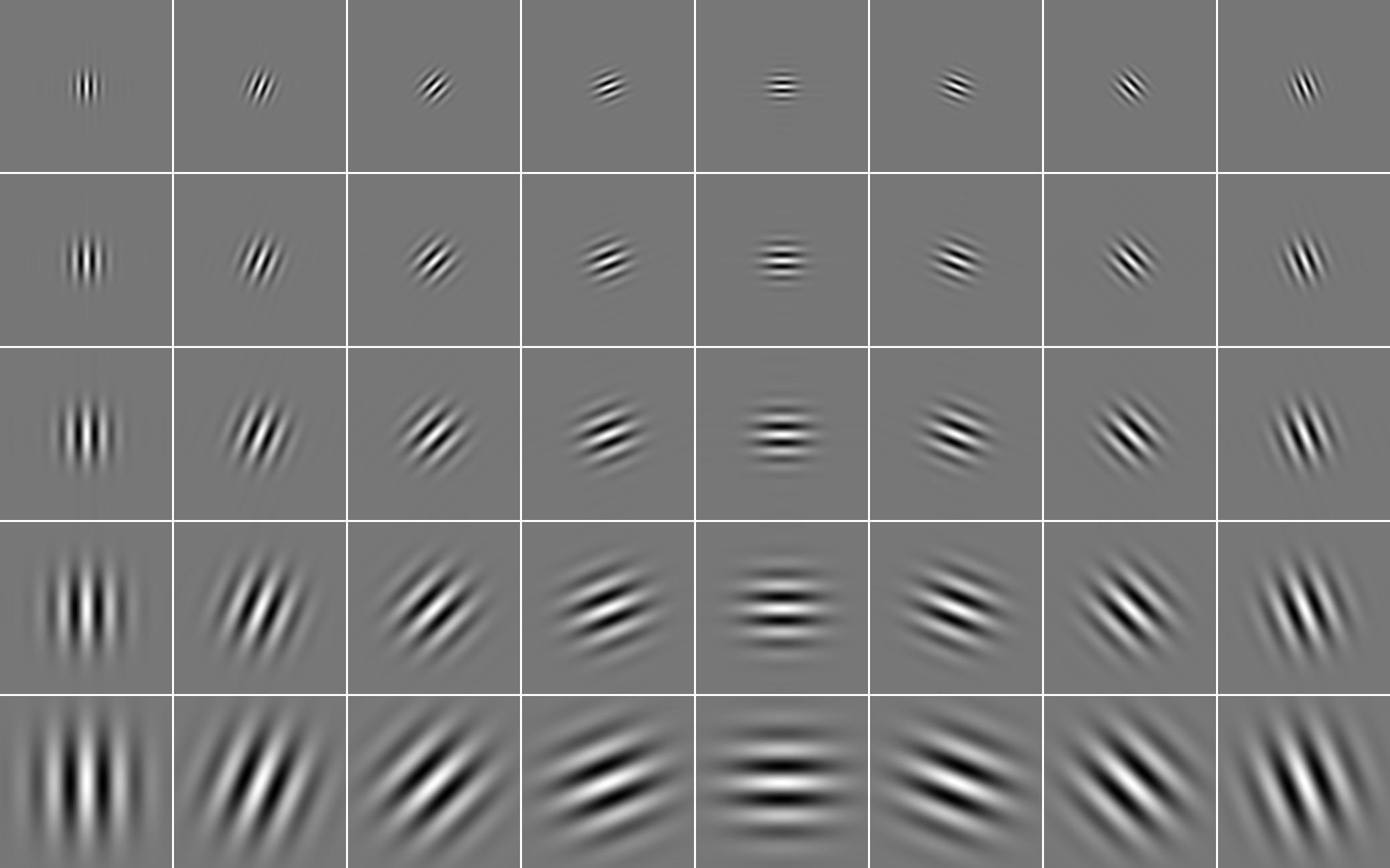}
\caption{Example of the 2-D complex Gabor filter bank with 40
coefficients (5 frequencies and 8 orientations). The coefficients
are computed by $\omega=2^{-(i+2)/2}$ ($i=0, ..., 4$),
$\theta=k\pi/8$ ($k=0, ..., 7$) and
$\sigma=2\pi/\omega$\cite{FR-Gabor97}.} \label{fig:GFB}
\end{figure}

These fast methods mentioned above primarily focus on reducing the
runtime of performing the 2-D complex Gabor filtering once at
specific orientation and frequency. However, in some computer vision
applications, it is often needed to compute the 2-D complex Gabor
filter bank consisting of the 2-D complex Gabor filtering outputs at
multiple orientations and frequencies. For instance, face
recognition approaches relying on the 2-D Gabor features usually
require performing the 2-D complex Gabor filtering at 8 orientations
and 5 frequencies (totally, 40 Gabor feature maps) to deal with
geometric variances \cite{FR-Gabor97, FR-NN03, FR-FG08,
wifs/GangwarJ15, FR-IJMLC16}. Fig. \ref{fig:GFB} shows the example
of the filter kernels used in the 2-D complex Gabor filter bank. To
compute the complex Gabor filter bank, existing approaches simply
repeat the Gabor computation step for a given set of frequencies and
orientations without considering the computational redundancy that
exists in such repeated calculations.

In this paper, we propose a novel approach that efficiently compute
2-D complex \emph{Gabor filter bank} by reducing the computational
redundancy that arises when performing the 2-D complex Gabor
filtering at multiple orientations and frequencies. 
We first decompose the Gabor basis kernels by making use of the
trigonometric identities. This allows us to perform a fast
convolution with the Gaussian kernel in a separable manner for $x$
and $y$ dimensions. More importantly, our decomposition strategy
enables the substantial reduction of the computational complexity
when computing the 2-D complex Gabor filter bank at a set of orientations and frequencies. 
In our formulation, intermediate results of the 2-D complex Gabor
filtering computed at a specific orientation can be reused when
performing the 2-D complex Gabor filtering at a symmetric orientation. 
This is particularly useful in some applications where the 2-D
complex Gabor filtering outputs at various orientations and
frequencies are needed to cope with geometric variations
\cite{FR-Gabor97, FR-NN03, tip/KamarainenKK06, FR-FG08, FR-IJMLC16,
wifs/GangwarJ15, tip/LiDZ15}. We will show that our method reduces
the computational complexity when compared to state-of-the-art
methods \cite{tsp/YoungVG02, tip/BernardinoS06}, while maintaining
the similar filtering quality.


Additionally, we present a method that efficiently computes 2-D
\emph{localized} sliding discrete Fourier transform (SDFT) using the
Gaussian kernel at the transform window by extending the proposed
kernel decomposition technique. In literature, the 2-D SDFT usually
performs the transform at an image patch within the transform window
by shifting the window one pixel at a time in either horizontal or
vertical directions. Numerous methods have been proposed for the
fast computation of the 2-D SDFT \cite{SDFT-SPM03, SDFT-SPM04,
SDFT-TIP15}. For instance, the relation between two successive 2-D
DFT outputs is first derived using the circular shift property
\cite{SDFT-TIP15}. Using this relation, the 2-D DFT output at the
current window is efficiently updated by linearly combining the 2-D
DFT output at the previous window and one 1-D DFT result only. Note
that all these methods use a box kernel 
within the sliding transform window and the circular shift property
holds only when the box kernel is employed. Therefore, applying the
existing 2-D SDFT methods \cite{SDFT-SPM03, SDFT-SPM04, SDFT-TIP15}
are infeasible in the case of calculating the \emph{localized} DFT
outputs with the Gaussian kernel.

It is generally known that the good spatial localization of the
Gabor filter mainly benefits from the use of the Gaussian kernel
which determines an weight based on a spatial distance. We will
present that the fast computation method of the 2-D \emph{localized}
SDFT using the Gaussian kernel, which lends the spatial localization
ability as in the Gabor filter, is also feasible using our
decomposition strategy. It should be noted that existing fast 2-D
complex Gabor filters \cite{tsp/YoungVG02, tip/BernardinoS06} can be
readily used to compute the 2-D localized SDFT, but a direct
application of theses methods disregards the computational
redundancy that exists on the repeated calculation of 2-D DFT
outputs at multiple frequencies as in the 2-D complex Gabor filter
bank.
We will show that our method outperforms existing approaches
\cite{tsp/YoungVG02, tip/BernardinoS06} in terms of
computational complexity. 


To sum up, our contributions can be summarized as follows.
\begin{itemize}

\item A new method is presented for efficiently computing the 2-D complex \emph{Gabor filter
bank} at a set of orientations and frequencies. We show that our
method runs faster than existing approaches.

\item The proposed method is extended into the 2-D \emph{localized} SDFT,
demonstrating a substantial runtime gain over existing approaches.

\item Extensive comparison with state-of-the-arts approaches is given in both
analytic and experimental manners.
\end{itemize}

The rest of this paper is organized as follows. In Section II, we
present the proposed method for fast computation of the 2-D complex
Gabor filter bank. In Section III, we present how the proposed
approach is extended to accelerate the 2-D localized SDFT. Section
IV presents experimental results including runtime and filtering
quality comparison with state-of-the-arts methods. Section V
concludes this paper with some remarks.

\section{Fast 2-D Complex Gabor Filter}
This section presents a new method that efficiently computes the 2-D
complex Gabor filter bank consisting of the 2-D complex Gabor
filtering outputs at multiple orientations and frequencies. We first
explain the Gabor kernel decomposition method to reduce the
complexity of 2-D complex Gabor filtering, and then show how the
decomposition method can be used for fast computation of 2-D complex
Gabor filter bank. 

For specific orientation $\theta$ and frequency $\omega$, the 2-D
complex Gabor filtering output $F_{\omega,\theta,\sigma}$ of a 2-D
image $f$ of $H\times W$ can be written as
\begin{equation}
\label{eq:original_2d}
\begin{split}
F_{\omega,\theta,\sigma}(x,y) = \sum\limits_{l,k} f (k,l)
C_{\omega,\theta}(x-k,y-l) {G_\sigma}(x-k,y-l)
\end{split}
\end{equation}
where $G_{\sigma}(x,y)$ is 2-D Gaussian function with zero mean and
the standard deviation of $\sigma$. Here, an isotropic Gaussian
kernel that has the same standard deviation for both $x$ and $y$
dimensions is used as in existing work \cite{tsp/YoungVG02,
tip/BernardinoS06}, i.e., $G_\sigma(x,y)=S_\sigma(x) S_\sigma(y)$.
The complex exponential function $C_{\omega,\theta}(x,y)$ for
orientation $\theta$ and frequency $\omega=2\pi/\lambda$, where
$\lambda$ represents wavelength, is defined as
\begin{equation}
\label{eq:exp}C_{\omega,\theta}(x,y) = e^{i\omega\left(x\cos\theta +
y\sin\theta\right)}.
\end{equation}

\noindent This is decomposed as $C_{\omega,\theta}(x,y) =
H_{\omega,\theta}(x)V_{\omega,\theta}(y)$ with $H_{\omega,\theta}(x)
= e^{i\omega x\cos\theta}$, $V_{\omega,\theta}(y)=e^{i\omega
y\sin\theta}$.


\subsection{Kernel Decomposition}
Since $G_\sigma(x,y)$ and $C_{\omega,\theta}(x,y)$ are separable for
$x$ and $y$ dimensions, \eqref{eq:original_2d} can be rewritten as
\begin{equation}\label{eq:fh}
J_{\omega,\theta,\sigma}(x,y)=\sum\limits_{k} f (k,y)H_{\omega,\theta}(x-k){S_\sigma }(x-k),\\
\end{equation}
\begin{equation}\label{eq:original_2d_sep}
F_{\omega,\theta,\sigma}(x,y) = \sum\limits_{l}
J_{\omega,\theta,\sigma}(x,l) V_{\omega,\theta}(y-l){S_\sigma}(y-l),
\end{equation}

\noindent $J_{\omega,\theta,\sigma}$ is first computed by performing
1-D horizontal Gabor filtering and this is then used in 1-D vertical
filtering for obtaining the final Gabor output
$F_{\omega,\theta,\sigma}$.

\subsubsection{Horizontal 1-D Gabor Filtering}
We first present the efficient computation of
$J_{\sigma,\omega,\theta}$ in \eqref{eq:fh} based on the basis
decomposition using the trigonometric identities. We explain the
real part of $J_{\sigma,\omega,\theta}$ only, as its imaginary
counterpart can be decomposed in a similar manner. For the sake of
simplicity, we define $\omega^c_{\theta}=\omega \cos\theta$ and
$\omega^s_{\theta}=\omega \sin\theta$. We also omit $y$ in the
computation of $J_{\sigma,\omega,\theta}$ and $f$ as the 1-D
operation is repeated for $y=1,...,H$. Using the trigonometric
identity $\cos(a-b)=\cos a\cos b+\sin a \sin b$, we can simply
decompose \eqref{eq:fh} into two terms as
\begin{flalign}\label{eq:pullout_real}
\mathcal{R}\{J_{\omega,\theta,\sigma}(x)\} &= \cos (\omega^c_{\theta} x)\sum\limits_{k} {f_c(k){S_\sigma }(x-k)} \nonumber\\
&+\sin (\omega^c_{\theta} x)\sum\limits_{k} {f_s(k){S_\sigma
}(x-k)},
\end{flalign}

\noindent where $f_c(k) = f(k)\cos (\omega^c_{\theta} k)$ and
$f_s(k) = f(k)\sin (\omega^c_{\theta} k)$. $\mathcal{R}(F)$
represents the real part of $F$. Then, \eqref{eq:pullout_real} can
be simply computed by applying the Gaussian smoothing to two
modulated signals $f_c$ and $f_s$, respectively. The imagery
counterpart $\mathcal{I}\{F\}$ can be expressed similarly as
\begin{flalign}\label{eq:pullout_imag}
\mathcal{I}\{J_{\omega,\theta,\sigma }(x)\} &= -\cos (\omega^c_{\theta} x)\sum\limits_{k} {f_s(k){S_\sigma }(x-k)} \nonumber\\
&+\sin (\omega^c_{\theta} x)\sum\limits_{k} {f_c(k){S_\sigma
}(x-k)}.
\end{flalign}

Interestingly, both real and imagery parts of $F$ contains the
Gaussian convolution with $f_c$ and $f_s$, thus requiring only two
1-D Gaussian smoothing in computing \eqref{eq:pullout_real} and
\eqref{eq:pullout_imag}. Many methods have been proposed to perform
fast Gaussian filtering \cite{sigpro/YoungV95,icpr/VlietYV98}, where
the computational complexity per pixel is independent of the
smoothing parameter $\sigma$. Here, we adopted the recursive
Gaussian filtering proposed by Young and Vliet
\cite{icpr/VlietYV98}.

\subsubsection{Vertical 1-D Gabor Filtering}
After $J(x,y)$ is computed using \eqref{eq:pullout_real} and
\eqref{eq:pullout_imag} for all $y=1,..,H$, we perform the 1-D Gabor
filtering on the vertical direction using
\eqref{eq:original_2d_sep}. Note that the input signal $J$ in
\eqref{eq:original_2d_sep} is complex, different from the real input
signal $f$ in \eqref{eq:fh}. Using the trigonometric identity, we
decompose the real and imagery parts of $F$ in
\eqref{eq:original_2d_sep} as follows:
\begin{flalign}\label{eq:2d_complex_real}
{\cal R}\{ {F_{\omega,\theta,\sigma}(x,y)}\} &= \cos (\omega_{\theta}^sy) \left( { f'_{cr}(x,y) + f'_{si}(x,y) } \right) \nonumber\\
&+ \sin (\omega _{\theta }^sy) \left( {f'_{sr}(x,y) - f'_{ci}(x,y)}
\right),
\end{flalign}
\begin{flalign}\label{eq:2d_complex_imag}
{\cal I}\{ {F_{\omega ,\theta, \sigma}(x,y)}\} &= \sin (\omega_{\theta }^sy) \left( {f'_{cr}(x,y) + f'_{si}(x,y)} \right) \nonumber\\
& - \cos (\omega _{\theta }^sy) \left( {f'_{sr}(x,y) - f'_{ci}(x,y)}
\right).
\end{flalign}


\noindent Here, $f'_{cr}$, $f'_{sr}$, $f'_{ci}$, and $f'_{si}$ are
filtering results convolved with 1-D Gaussian kernel $S_\sigma$ as
follows:
\begin{equation}\label{eq:modulated_signal_conv}
\begin{array}{l}
f'_{cr}(x,y)+f'_{si}(x,y) = \sum\limits_l {(f_{cr}(x,l)+f_{si}(x,l)){S_\sigma }(y - l)},\\
f'_{sr}(x,y)-f'_{ci}(x,y) = \sum\limits_l
{(f_{sr}(x,l)-f_{ci}(x,l)){S_\sigma }(y - l)},
\end{array}
\end{equation}


\noindent where the modulated signals $f_{cr}$, $f_{sr}$, $f_{ci}$,
and $f_{si}$ are defined as
\begin{equation}\label{eq:modulated_signal}
\begin{array}{l}
f_{cr}(x,y)=\mathcal{R}\{J_{\omega,\theta,\sigma}(x,y)\}\cos(\omega^s_{\theta}y),\\
f_{sr}(x,y)=\mathcal{R}\{J_{\omega,\theta,\sigma}(x,y)\}\sin(\omega^s_{\theta}y),\\
f_{ci}(x,y)=\mathcal{I}\{J_{\omega,\theta,\sigma}(x,y)\}\cos(\omega^s_{\theta}y),\\
f_{si}(x,y)=\mathcal{I}\{J_{\omega,\theta,\sigma}(x,y)\}\sin(\omega^s_{\theta}y).
\end{array}
\end{equation}

\noindent Like the horizontal filtering, two 1-D Gaussian
convolutions are required in \eqref{eq:2d_complex_real} and
\eqref{eq:2d_complex_imag}, i.e., $f'_{cr}(x,l) + f'_{si}(x,l)$ and
$f'_{sr}(x,l) - f'_{ci}(x,l)$.



In short, decomposing the complex exponential basis function
$C_{\omega,\theta}$ enables us to apply fast Gaussian filtering
\cite{sigpro/YoungV95,icpr/VlietYV98}. Though the fast Gaussian
filter was used for implementing fast recursive Gabor filtering in
\cite{tsp/YoungVG02}, our method relying on the trigonometric
identity and separable implementation for $x$ and $y$ dimensions
results in a lighter computational cost than
the state-of-the-arts method \cite{tsp/YoungVG02}. 
More importantly, we will show this decomposition further reduces
the computational complexity when computing the 2-D complex Gabor
filter bank. 

\subsection{Fast Computation of 2-D Complex Gabor Filter Bank}
Several computer vision applications often require computing the 2-D
complex Gabor filter bank consisting of a set of 2-D complex Gabor
filtering outputs at multiple frequencies and orientations. For
instance, in order to deal with geometric variances, some face
recognition approaches use the 2-D complex Gabor filtering outputs
at 8 orientations and 5 frequencies (see Fig. \ref{fig:GFB}) as
feature descriptors \cite{FR-Gabor97, FR-NN03, FR-FG08,
wifs/GangwarJ15, FR-IJMLC16}. To compute the 2-D complex Gabor
filter bank, existing approaches repeatedly perform the 2-D complex
Gabor filtering for a given set of frequencies and orientations,
disregarding the computational redundancy that exists in such
repeated calculations.

In this section, we present a new method that efficiently computes
the 2-D complex Gabor filter bank. Without the loss of generality,
it is assumed that the standard deviation $\sigma$ of the Gaussian
kernel is fixed. For a specific frequency $\omega$, we aim at
computing the 2-D complex Gabor filter bank at $N$ orientations
$\{\frac{\pi k}{N}|k=0,...,N-1\}$. Here, $N$ is typically used as an
even number. For the simplicity of notation, we omit $\omega$ and
$\sigma$ in all equations. Let us assume that $F_\theta$ in
\eqref{eq:original_2d_sep} is computed using the proposed kernel
decomposition technique and its intermediate results are stored. We
then compute $F_{\pi-\theta}$ by recycling these intermediate
results. The separable form of $F_{\pi-\theta}$ can be written as

\begin{equation}\label{eq:fh2}
J_{\pi-\theta}(x,y)=\sum\limits_{k} f
(k,y)H_{\omega,\pi-\theta}(x-k){S_\sigma }(x-k).
\end{equation}
\begin{equation}\label{eq:pi-theta_2d_sep}
F_{\pi-\theta}(x,y) = \sum\limits_{l} J_{\pi-\theta}(x,l)
V_{\omega,\pi-\theta}(y-l){S_\sigma }(y-l)
\end{equation}

Using $H_{\omega,\pi-\theta}(x)=H^{*}_{\omega,\theta}(x)$, where $*$
denotes complex conjugation, \eqref{eq:fh2} can be rewritten as
follows:
\begin{equation}\label{eq:fh3}
J_{\pi-\theta}(x,y)=J^{*}_{\theta}(x,y).
\end{equation}
The horizontal 1-D Gabor filtering result $J_{\pi-\theta}$ is
complex conjugate to $J_{\theta}$. 
Using $V_{\omega,\pi-\theta}(x)=V_{\omega,\theta}(x)$, the vertical
1-D Gabor filtering in \eqref{eq:pi-theta_2d_sep} is then expressed
as
\begin{flalign} \label{eq:pi-theta_2d_sep2}
F_{\pi-\theta}(x,y) &= \sum\limits_{l} J^{*}_{\theta}(x,l)
V_{\omega,\theta}(y-l){S_\sigma }(y-l).
\end{flalign}

\noindent 
$F_{\pi-\theta}$ is obtained by applying the vertical 1-D Gabor
filtering to the complex conjugate signal $J^*_\theta$. Using
\eqref{eq:2d_complex_real} and \eqref{eq:2d_complex_imag}, the
following equations are derived:

\begin{flalign}\label{eq:pi-theta_real}
{\cal R}\{ {F_{\pi-\theta}(x,y)}\} &= \cos (\omega _{\theta }^sy) \left( {f'_{cr}(x,y) - f'_{si}(x,y) } \right) \nonumber\\
&+ \sin (\omega _{\theta }^sy) \left( {f'_{sr}(x,y) + f'_{ci}(x,y) }
\right),
\end{flalign}
\begin{flalign}\label{eq:pi-theta_imag}
{\cal I}\{ {F_{\pi-\theta}(x,y)}\} &= \sin (\omega _{\theta }^sy) \left( {f'_{cr}(x,y) - f'_{si}(x,y)} \right) \nonumber\\
&- \cos (\omega _{\theta }^sy) \left( {f'_{sr}(x,y) + f'_{ci}(x,y)}
\right)
\end{flalign}

\noindent The vertical filtering also requires two Gaussian
convolutions. Algorithm \ref{algo:pseudo_Gabor} summarizes the
proposed method for computing the 2-D complex Gabor filter bank.
When a set of frequencies $\Omega$ and orientations $\Theta$ are
given, we compute the 2-D complex Gabor filtering results at
$\theta_k$ ($k=0,...,N-1$) with the frequency $\omega_i$ being
fixed. Different from existing approaches \cite{tsp/YoungVG02,
tip/BernardinoS06} repeatedly applying the 2-D complex Gabor filter
at all orientations, we consider the computational redundancy that
exists on such repeated calculations to reduce the
runtime. 
We will demonstrate our method runs faster than existing fast Gabor
filters \cite{tsp/YoungVG02,tip/BernardinoS06} through both
experimental and analytic comparisons.

\begin{algorithm}
\caption{Pseudo code of 2-D complex Gabor filter
bank}\label{algo:pseudo_Gabor}
\begin{algorithmic}[1]
\State\textbf{Input}: input image $f$ ($H\times W$),\\
\quad\quad\; a set of $O$ scales $\Sigma =\{\sigma_i|i=0,...,O-1\}$, \\
\quad\quad\; a set of $O$ frequencies $\Omega =\{\omega_i|i=0,...,O-1\}$, \\
\quad\quad\; a set of $N$ orientations $\Theta=\{\theta_k|
k=0,...,N-1\}$

\State\textbf{Output}: 2-D complex Gabor filter outputs for $\Omega$
and $\Theta$  \vspace{8pt}


\For{$i = 0, ..., O-1$}  \Comment{For all frequencies}

\State $\sigma_i = 2\pi/\omega_i, N_h=\lfloor N/2 \rfloor$ 
\vspace{8pt}

\For{$k = 0, ..., N_h$}  \Comment{For half of all orientations}
\State $\theta_k = \pi k/N$

\For{$y = 1, ..., H$}
\State Perform 1-D Gaussian filtering of $f_c, f_s$
\State Compute $J_{\omega_i,\theta_k,\sigma_i}(x,y)$ for all $x$ in \eqref{eq:pullout_real} and \eqref{eq:pullout_imag}
\EndFor

\For{$x = 1,..., W$}
\State Perform 1-D Gaussian filtering of $f_{cr} + f_{si}$,

\quad\quad\; $f_{sr} - f_{ci}$ in \eqref{eq:2d_complex_real} and \eqref{eq:2d_complex_imag}
\State Compute $F_{\omega_i,\theta_k,\sigma_i}(x,y)$ for all $y$
\EndFor

\EndFor
\vspace{8pt}

\For{$k = N_h+1, ..., N-1$}  \Comment{For remaining ori.} \State
$\theta_k = \pi k/N$

\State $J_{\omega_i,\theta_k,\sigma_i}(x,y) = J^*_{\omega_i,\pi-\theta_k,\sigma_i}(x,y)$ for all $x$ and $y$.

\For{$x = 1,..., W$} \State Perform 1-D Gaussian filtering of $f_{cr} - f_{si}$,

\quad\quad\; $f_{sr} + f_{ci}$ in \eqref{eq:2d_complex_real}  and \eqref{eq:2d_complex_imag}
\State Compute $F_{\omega_i,\theta_k,\sigma_i}(x,y)$ for all $y$
\EndFor

\EndFor \vspace{8pt}

\EndFor

\end{algorithmic}
\end{algorithm}




\section{Localized Sliding DFT}
It is known that the Gabor filter offers the good spatial
localization thanks to the Gaussian kernel that determines an weight
based on a spatial distance. Inspired by this, we present a new
method that efficiently compute the 2-D \emph{localized} SDFT using
the proposed kernel decomposition technique. Different from the
existing 2-D SDFT approaches \cite{SDFT-SPM03, SDFT-SPM04,
SDFT-TIP15} using the box kernel, we use the Gaussian kernel when
computing the DFT at the sliding window as in the Gabor filter. It
should be noted that applying the existing 2-D SDFT approaches
\cite{SDFT-SPM03, SDFT-SPM04, SDFT-TIP15} are infeasible in the case
of calculating the DFT outputs with the Gaussian kernel.

\subsection{Kernel decomposition in 2-D localized SDFT}
When the sliding window of $M\times M$ is used, we set the standard
deviation $\sigma$ of the Gaussian kernel by considering a cut-off
range, e.g., $\lfloor M/2 \rfloor =3\sigma$. We denote
$F_{u,v}(x,y)$ by the $(u,v)^{th}$ bin of the $M\times M$ DFT at
$(x,y)$ of 2-D image $f$. The 2-D localized SDFT with the Gaussian
kernel can be written as
\begin{flalign} \label{eq:sdft-ori}
 F_{u,v} &(x,y) = \nonumber\\
 &{\sum\limits_{n,m} {f(m,n)C_{u,v} (\widehat{x}-m,\widehat{y}-n)G_\sigma(x-m,y-n)} }
\end{flalign}

\noindent where $\widehat{x}=x-\frac{M}{2}$ and
$\widehat{y}=y-\frac{M}{2}$. For $u,v=0,...,M-1$, the complex
exponential function $C_{u,v} (m,n)$ at the $(u,v)^{th}$ frequency
is defined as
\begin{equation} \label{eq:sdft-exp}
C_{u,v} (x,y) = e^{ i ( \omega_0ux + \omega_0vy)},
\end{equation}

\noindent where $\omega_0=\frac{2\pi}{M}$. Note that in
\eqref{eq:sdft-ori} and \eqref{eq:sdft-exp}, slightly different
notations than the conventional SDFT methods \cite{SDFT-SPM03,
SDFT-SPM04, SDFT-TIP15} are used to keep them consistent with the
Gabor filter of \eqref{eq:original_2d}. When $G_\sigma(x,y)=1$,
\eqref{eq:sdft-ori} becomes identical to that of the conventional
SDFT methods \cite{SDFT-SPM03, SDFT-SPM04, SDFT-TIP15}. The Gaussian
window of $M\times M$ is used here, but the 2-D localized SDFT using
$M_y\times M_x$ window ($M_y\neq M_x$) is also easily derived.

Using the separable property of
$G_\sigma(x,y)=S_\sigma(x)S_\sigma(y)$ and $C_{u,v}(x,y)=H_u(x)
V_v(y)$, \eqref{eq:sdft-ori} can be written as
\begin{equation} \label{eq:sdft-h}
 J_{u} (x,y) = \sum\limits_{m} {f(m,y)H_u (\widehat{x}-m)S_\sigma
 (x-m)},
\end{equation}
\begin{equation} \label{eq:sdft-v}
 F_{u,v} (x,y) = \sum\limits_{n} {J_{u} (x,n)V_v (\widehat{y}-n)S_\sigma (y-n)}.
\end{equation}

\noindent Using the kernel decomposition, the 1-D horizontal
localized SDFT is performed as follows:
\begin{flalign}\label{eq:sdft-h-real}
\mathcal{R}\{J_{u}(x)\} &= \cos (\omega_0 u\widehat{x})\sum\limits_{m} {f_c(m){S_\sigma }(x-m)} \nonumber\\
&+\sin (\omega_0 u\widehat{x})\sum\limits_{m} {f_s(m){S_\sigma
}(x-m)},
\end{flalign}
\begin{flalign}\label{eq:sdft-h-imag}
\mathcal{I}\{J_{u}(x)\} = &-\cos (\omega_0 u\widehat{x})\sum\limits_{m} {f_s(m){S_\sigma }(x-m)} \nonumber\\
&+\sin (\omega_0 u\widehat{x})\sum\limits_{m} {f_c(m){S_\sigma
}(x-m)},
\end{flalign}

\noindent where $f_c(m)=f(m)cos(\omega_0 um)$,
$f_s(m)=f(m)sin(\omega_0 um)$.

The vertical 1-D localized SDFT is performed similar to the Gabor
filter:
\begin{flalign}\label{eq:sdft-v-real}
{\cal R}\{ {F_{u,v}(x,y)}\} &= \cos (\omega_0 v\widehat{y}) \left( { f'_{cr}(x,y) + f'_{si}(x,y) } \right) \nonumber\\
&+ \sin (\omega_0 v\widehat{y}) \left( {f'_{sr}(x,y) - f'_{ci}(x,y)}
\right),
\end{flalign}
\begin{flalign}\label{eq:sdft-v-imag}
{\cal I}\{ {F_{u,v}(x,y)}\} &= \sin (\omega_0 v\widehat{y}) \left( {f'_{cr}(x,y) + f'_{si}(x,y)} \right) \nonumber\\
& - \cos (\omega_0 v\widehat{y}) \left( {f'_{sr}(x,y) -
f'_{ci}(x,y)} \right),
\end{flalign}
where $f'_{cr}+f'_{si}$ and $f'_{sr} - f'_{ci}$ are defined in a
manner similar to \eqref{eq:modulated_signal_conv}.

\subsection{Exploring Computational Redundancy on $(u,v)$}
The 2-D localized SDFT requires computing a set of DFT outputs for
$u,v=0,...,M-1$, similar to the 2-D complex Gabor filter bank.
Considering the conjugate symmetry property of the DFT
($F_{M-u,M-v}=F^*_{u,v}$), we compute the DFT outputs $F_{u,v}$ only
for $u=0,...,M-1$ and $v=0,...,\lfloor M/2\rfloor$, and then simply
compute remaining DFT outputs (for $u=0,...,M-1$ and $v=\lfloor
M/2\rfloor +1,...,M-1$) by using the complex conjugation. Thus, we
focus on the computation of the 2-D SDFT for $u=0,...,M-1$ and
$v=0,...,\lfloor M/2 \rfloor$.

Let us consider how to compute $F_{M-u,v}$ using intermediate
results of $F_{u,v}$. Similar to the Gabor filter bank, the 1-D DFT
$J_{M-u}$ is complex conjugate to $J_{u}$ as follows:
\begin{flalign} \label{eq:sdft-h2}
 J_{M-u} (x,y) &= \sum\limits_{m} {f(m,y)H_{M-u} (\widehat{x}-m)S_\sigma (x-m)}, \nonumber\\
 &= \sum\limits_{m} {f(m,y)H^*_{u} (\widehat{x}-m)S_\sigma (x-m)}, \nonumber\\
 &= J^*_u (x,y)
\end{flalign}
\noindent The 1-D vertical SDFT result $F_{M-u,v}$ is then obtained
as
\begin{flalign} \label{eq:sdft-v2}
 F_{M-u,v} (x,y) = \sum\limits_{n} {J^*_{u} (x,n)V_v (\widehat{y}-n)S_\sigma (y-n)}.
\end{flalign}

\noindent As in the Gabor filter bank, \eqref{eq:sdft-v2} can be
computed by performing the 1-D vertical Gaussian filtering twice.

\begin{figure}[t]
 \renewcommand{\thesubfigure}{}
  \centering
  \subfigure [(a) Log polar grid]
{\includegraphics[width=0.23\textwidth]{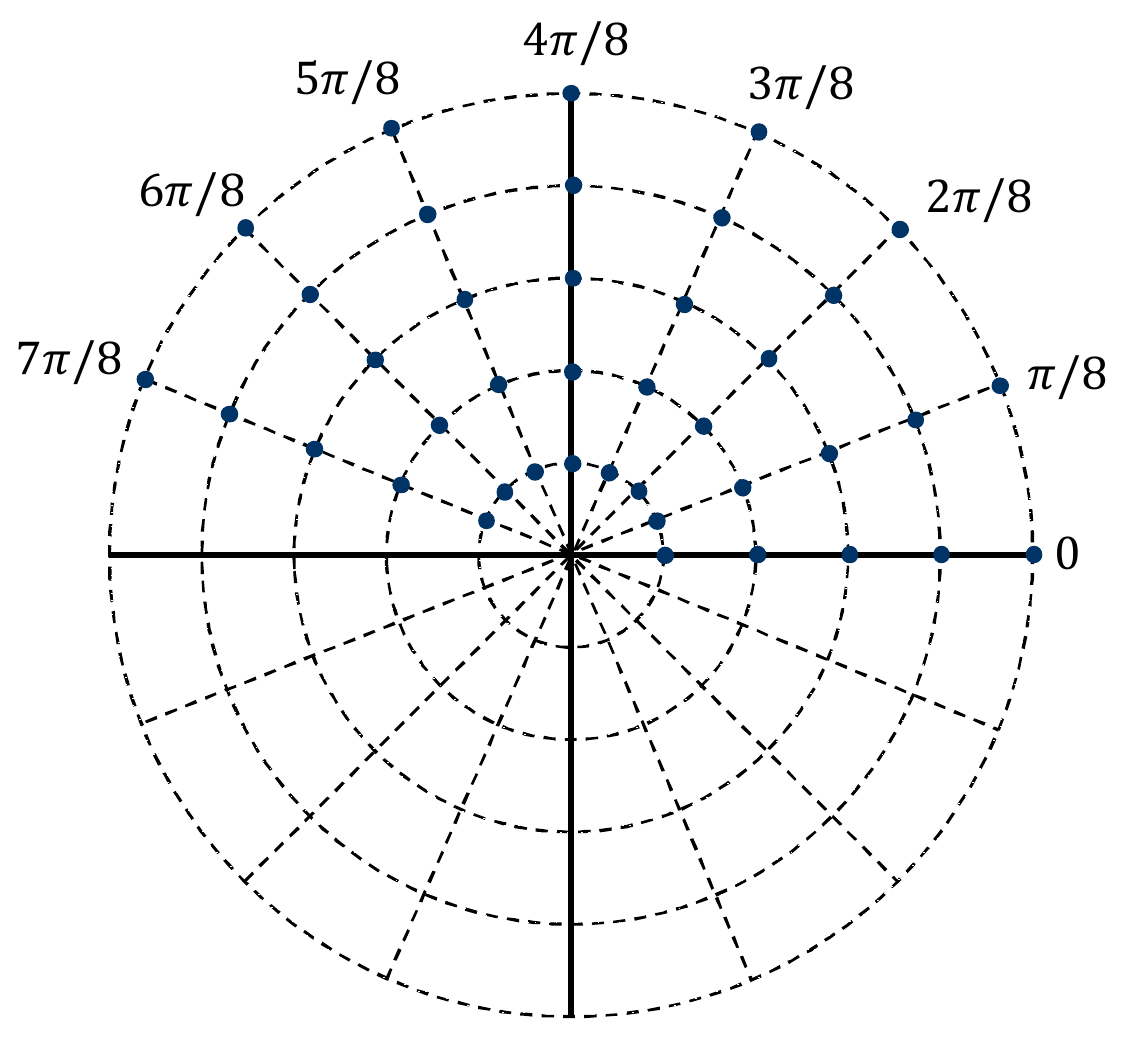}}
\hspace{0.2cm}
  \subfigure [(b) Rectangular grid]
{\includegraphics[width=0.23\textwidth]{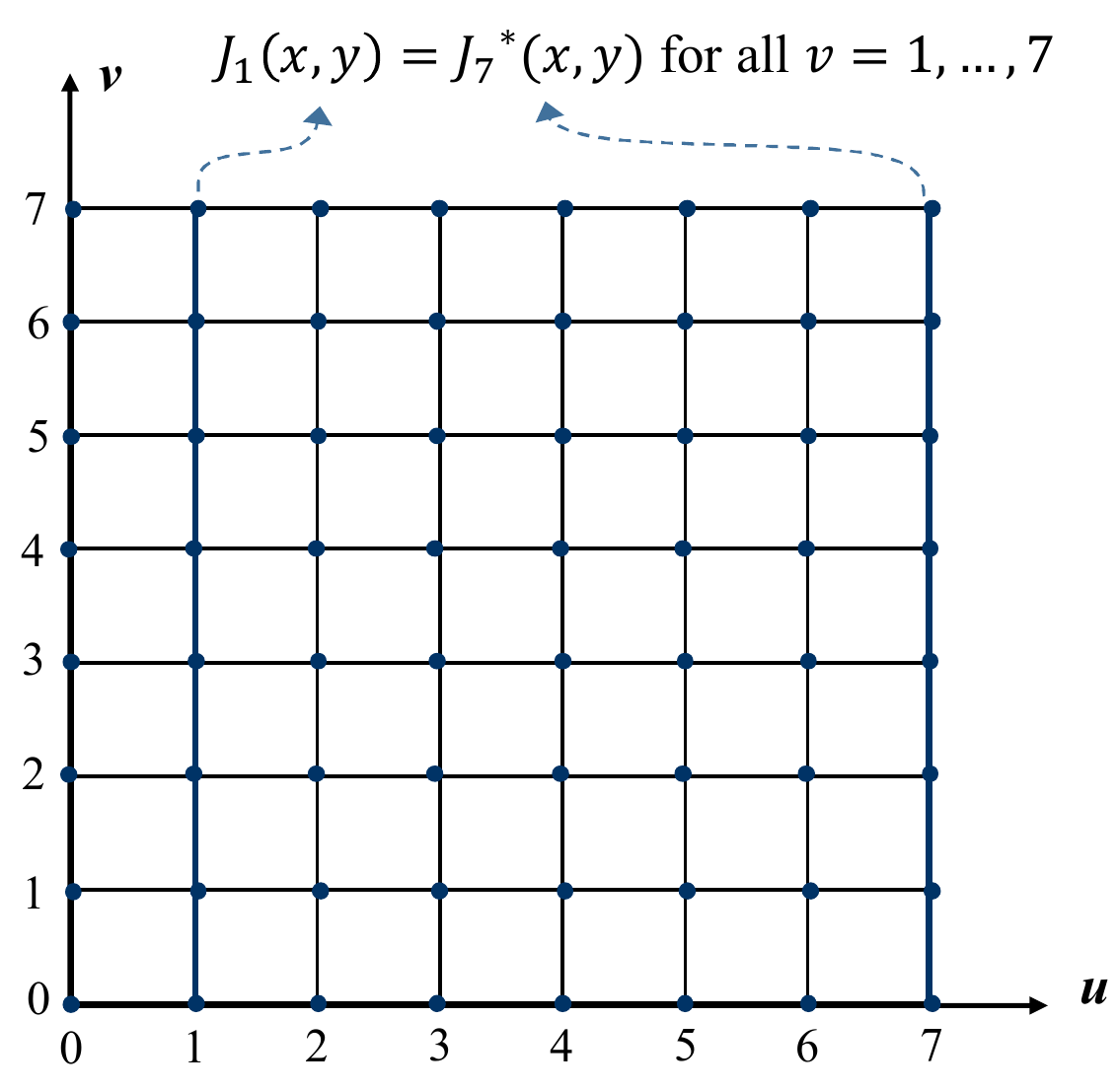}}
\caption{Log polar grid of the 2-D complex Gabor filter and the
rectangular grid of the 2-D SDFT. (a) 5 frequencies and 8
orientations, (b) $8\times8$ window ($M=8$). In the log polar grid,
two 1-D horizontal Gabor outputs are complex conjugate, i.e.,
$J_{\omega, \theta, \sigma}=J^*_{\omega, \pi-\theta, \sigma}$, when
$\omega$ is fixed. In the 2-D SDFT, $J_u=J_{M-u}$ holds for
$v=0,...,M-1$. These intermediate results can be reused in the
computation of Gabor filter bank and the 2-D localized SDFT.}
\label{fig:grid_of_Gabor_SDFT} \vspace{0.5cm}
\end{figure}

Fig. \ref{fig:grid_of_Gabor_SDFT} visualizes the log polar grid of
the 2-D complex Gabor filter and the regular grid of the 2-D SDFT.
There exists an additional computational redundancy when performing
the 2-D SDFT on the regular grid. Specifically, for a specific $u$,
the 1-D horizontal filtering results $J_u(x,y)$ remain unchanged for
$v=0,...,\lfloor M/2\rfloor$. These intermediate results can be used
as inputs for the 1-D vertical localized SDFT for $v=0,...,\lfloor
M/2\rfloor$.

\begin{algorithm}
\caption{Pseudo code of 2-D Localized SDFT}\label{algo:pseudo_SDFT}
\begin{algorithmic}[1]
\State\textbf{Input:} input image $f$ ($H\times W$), scale $\sigma$,
kernel size $M_y\times M_x$ ($M_y \geq M_x$)

\State\textbf{Output:} SDFT outputs at $u=0,...,M_x-1$ and $v=0,...,M_y-1$ \vspace{8pt}

\State{$M_{xh}=\lfloor M_x/2 \rfloor$},
{$M_{yh}=\lfloor M_y/2 \rfloor$}

\For {$u = 0,..., M_{xh}$}
\For {$y=1,...,H$}    \Comment{1-D horizontal SDFT}
\State {Perform 1-D Gaussian filtering of $f_c, f_s$}

\quad in \eqref{eq:sdft-h-real} and \eqref{eq:sdft-h-imag}. \State
Compute $J_{u}(x,y)$ for all $x$. \EndFor \EndFor

\For{$u = M_{xh}+1,..., M_x-1$} \State $J_{M_x-u}(x,y) = J^*_{u}(x,y)$
for all $x$ and $y$. \EndFor \vspace{8pt}

\For{$u = 0, ..., M_x-1$, $v = 0, ..., M_{yh}$}
\For{$x=1,...,W$}   \Comment{1-D vertical SDFT}
\State Perform 1-D Gaussian filtering of $f_{cr} + f_{si}$,

\quad $f_{sr} - f_{ci}$ in \eqref{eq:sdft-v-real} and \eqref{eq:sdft-v-imag}.
\State Compute $F_{u,v}(x,y)$ for all $y$. 
\EndFor \EndFor \vspace{8pt}

\For{$u = 0, ..., M_x-1$, $v =M_{yh}+1, ..., M_y-1$} \State
$F_{u,v}(x,y)=F^*_{M_x-u,M_y-v}(x,y)$ \EndFor
\end{algorithmic}
\end{algorithm}

Algorithm \ref{algo:pseudo_SDFT} shows the overall process of
computing the 2-D localized SDFT. Here, we explain the method with a
non-square window of $M_y\times M_x$ ($M_y \geq M_x$) for a
generalized description. This can be simply modified when $M_y <
M_x$. Note that when $M_y\geq M_x$, a horizontal filtering (line
$4-9$ of Algorithm \ref{algo:pseudo_SDFT}) should be performed first
and \emph{vice versa} in order to reduce the runtime. This filtering
order does not affect the computational complexity of the 1-D SDFT
in line $13-18$. In contrast, the 1-D SDFT in line $4-9$ is affected
when $M_y \neq M_x$, and thus we should perform the 1-D filtering
for $u=0,...,\lfloor M_x/2\rfloor$ on the horizontal direction in
line $4-9$ if $M_y \geq M_x$. The number of arithmetic operations is also reported in Table \ref{tb:SDFT-time2}. 

To obtain $M_y\times M_x$ DFT outputs at the sliding window of the
input image $f$ in Algorithm \ref{algo:pseudo_SDFT}, we first obtain
$J_u(x,y)$ for $u=0,...,\lfloor M_x/2\rfloor$ by using
\eqref{eq:sdft-h}, and then simply calculate $J_u(x,y)$ for
$u=\lfloor M_x/2\rfloor+1,...,M_x-1$ using \eqref{eq:sdft-h2}.
$J_{u}(x,y)$ computed once using the horizontal filtering can be
used to obtain $F_{u,v}(x,y)$ by performing the 1-D vertical
filtering. 
Thus, the horizontal filtering $J_u(x,y)$ is performed only for
$u=0,...,\lfloor M_x/2\rfloor$, while the vertical filtering
$F_{u,v}(x,y)$ is done for $u=0,...,M_x-1$ and $v=0,...,\lfloor
M_y/2\rfloor$.

%
%
%
%
%
%
%


\section{Experimental Results}
We compared the proposed method with state-of-the-arts methods
\cite{tsp/YoungVG02, tip/BernardinoS06} for fast Gabor filtering in
terms of both computational efficiency and filtering quality. 
For a fair comparison, we implemented the two methods
\cite{tsp/YoungVG02, tip/BernardinoS06} with a similar degree of
code optimization, and compared their runtime and filtering quality
through experiments. All the codes including our method will be
publicly available later for both the 2-D complex Gabor filter bank
and the 2-D localized SDFT.

\subsection{Computational Complexity Comparison}
\begin{table}
\begin{center}
\renewcommand{\arraystretch}{1.5}
\caption{Runtime comparison (millisecond) of the 2-D complex Gabor
filter bank. The recursive Gabor filter \cite{tsp/YoungVG02} and IIR
Gabor filter \cite{tip/BernardinoS06} are used for comparison. We
measured the runtime when computing the 2-D complex Gabor filter
bank for multiple orientations at a specific frequency. The set of
$N$ orientations $\Theta$ is defined as $\{\theta_k =
\frac{k\pi}{N}| k
= 1,....,N - 1 \}$. The input image is of $1024\times 1024$. 
} \label{tb:Gabor-time}

\begin{tabular}{>{\centering}p{5mm}|>{\centering}p{20mm}|>{\centering}p{15mm}|>{\centering}p{15mm}}
\hline $N$ & \textbf{Recursive Gabor fil.} \cite{tsp/YoungVG02} &
\textbf{IIR Gabor filter} \cite{tip/BernardinoS06} & \textbf{Ours}
\tabularnewline

\hline 8 & 608 & 500 & 359 \tabularnewline
\hline 14 & 1039 & 852 & 586 \tabularnewline
\hline 20 & 1518 & 1230 & 842 \tabularnewline
\hline 26 & 1972 & 1597 & 1079 \tabularnewline
\hline 32 & 2421 & 1971 & 1314 \tabularnewline

\hline
\end{tabular}
\end{center}
\end{table}

\begin{table}
\caption{Computational complexity comparison of the 2-D complex
Gabor filter bank. The recursive Gabor filter \cite{tsp/YoungVG02}
and fast IIR Gabor filter \cite{tip/BernardinoS06} are used for
comparison. Similar to Table \ref{tb:Gabor-time}, when computing the
2-D complex Gabor filter bank for $N$ orientations at a specific
frequency, we count the number of multiplications $R_M$ and
additions $R_A$ per pixel, respectively.} \label{tb:Gabor-time2}
\centering
\begin{tabular}{>{\centering}p{11mm}|>{\centering}p{11mm}|c|c|c|c|c|c}
\hline
    &   &
\multicolumn{6}{c}{The number of orientations $N$} \\
\cline{3-8}
Algorithm   &   Operation & 8 & 14 &  20 & 26 &    30   & $N$ \\
\hline \hline
\cite{tsp/YoungVG02}      & $R_{M}$ &416 & 728 & 1040 & 1352 & 1560 &   $52N$      \\
\cline{2-8}
          & $R_{A}$ & 376 & 658 & 940 & 1222 & 1410 &   $47N$          \\
\hline
\cite{tip/BernardinoS06}      & $R_{M}$ & 272 & 476 & 680 & 884 & 1020 &   $34N$      \\
\cline{2-8}
          & $R_{A}$ &208 & 364 & 520 & 676 & 780 &   $26N$     \\
\hline
Ours      & $R_{M}$ & 240 & 420 & 600 & 780 & 900  &   $30N$      \\
\cline{2-8}
          & $R_{A}$ & 176 & 308 & 440 & 572 & 660  &   $22N$       \\
\hline

\hline
\end{tabular}
\end{table}

We first compared the runtime when computing the 2-D complex Gabor
filter bank. As our method focuses on reducing the computational
redundancy on the repeated application of the 2-D complex Gabor
filter at multiple orientations, we compared only the runtime for
computing the 2-D complex Gabor filter bank. Additionally, the
runtime was analyzed by counting the number of arithmetic operations
such as addition and multiplication. The runtime of the 2-D
localized SDFT was also measured in both experimental and analytic
manners. The existing fast Gabor filters \cite{tsp/YoungVG02,
tip/BernardinoS06} can be applied to compute the 2-D localized SDFT
by computing the DFT outputs for all frequency bins. Conventional
2-D SDFT approaches using the box kernel \cite{SDFT-SPM03,
SDFT-SPM04, SDFT-TIP15} were not compared in the experiments, since
they are not capable of computing the 2-D
localized DFT outputs. 

Table \ref{tb:Gabor-time} compares the runtime in the computation of
the 2-D complex Gabor filter bank. As summarized in Algorithm
\ref{algo:pseudo_Gabor}, our method can be repeatedly applied to
each frequency. Thus, we measured the runtime in the computation of
the 2-D complex Gabor filter bank for $N$ orientations when a
specific frequency $\omega$ is given. The set of orientations
$\Theta$ is defined as
$\{\theta_k = \frac{k\pi}{N}| k = 0,....,N-1 \}$. 
In the existing fast Gabor filters \cite{tsp/YoungVG02,
tip/BernardinoS06}, there is no consideration of the computational
redundancy that occurs when computing the Gabor outputs at multiple
orientations. The fast Gabor filter using IIR approximation
\cite{tip/BernardinoS06} is computationally lighter than the
recursive Gabor filter \cite{tsp/YoungVG02}, but our method runs
faster than the two methods. In Table \ref{tb:Gabor-time2}, we
compare the number of arithmetic operations at $N$ orientations and
a single frequency $\omega$, in the manner similar to Table
\ref{tb:Gabor-time}. We count the number of multiplications $R_M$
and additions $R_A$ per pixel, respectively. Considering $R_M$ and
$R_A$ of the three approaches, the runtime results in Table
\ref{tb:Gabor-time} are in agreement. Again, the codes for the three
methods will be publicly available. 

\begin{table}
\begin{center}
\renewcommand{\arraystretch}{1.5}
\caption{Runtime comparison (millisecond) of the 2-D localized SDFT.
The window size for DFT is $M\times M$ where $\lfloor M/2\rfloor=3\sigma$ is set with
the standard deviation $\sigma$ of the Gaussian kernel. We also
compared with two existing methods \cite{tsp/YoungVG02,
tip/BernardinoS06} by repeatedly applying them when computing
$F_{u,v}$ for $u,v=0,...,M-1$. Note that the conjugate symmetry
property was used when measuring the runtime for all three methods.
Interestingly, the runtime gain becomes higher than that of the fast
Gabor filter bank in Table \ref{tb:Gabor-time}. The input image is
of $250\times 234$. For more details, refer to the text.}
\label{tb:SDFT-time}

\begin{tabular}{>{\centering}p{12mm}|>{\centering}p{20mm}|>{\centering}p{15mm}|>{\centering}p{15mm}}
\hline $M\times M$ & \textbf{Recursive Gabor fil.}
\cite{tsp/YoungVG02} & \textbf{IIR Gabor filter}
\cite{tip/BernardinoS06}  & \textbf{Ours} \tabularnewline \hline

$8\times8$ & 45 & 101 & 40 \tabularnewline \hline
$10\times10$ & 67 & 159& 57 \tabularnewline \hline
$12\times12$ & 94 & 228 & 75 \tabularnewline \hline
$14\times14$ & 127 & 317 & 101 \tabularnewline \hline
$16\times16$ & 163 & 421 & 125 \tabularnewline \hline

\end{tabular}
\end{center}
\vspace{-0.1cm}
\end{table}

\begin{table*}[t]
\caption{Computational complexity comparison of the 2-D localized
SDFT. Similar to Table \ref{tb:SDFT-time}, we compared with two
existing methods \cite{tsp/YoungVG02, tip/BernardinoS06}. The window
size of DFT is $M\times M$. We count the number of multiplications
$R_M$ and additions $R_A$ per pixel required to compute the 2-D DFT
$F_{u,v}$ for $u,v=0,...,M-1$. We also count $R_M$ and $R_A$ when a
non-square window of $M_y\times M_x$ ($M_y \neq M_x$) is used.}
\label{tb:SDFT-time2} \centering
\begin{tabular}{c|c|c|c|c|c|c|c|c}
\hline
    &   &
\multicolumn{7}{c}{Kernel size} \\
\cline{3-9}
Algorithm   &   Operation & $1\times1$ & $2\times2$ &  $4\times4$ & $8\times8$ &    $16\times16$    & $M_y\times M_x$ ($M_y \geq M_x$) & $M_y\times M_x$ ($M_y < M_x$) \\
\hline \hline

\textbf{Recursive Gabor fil.}
\cite{tsp/YoungVG02} & $R_{M}$ & 26 &   78  &   260  &   936 &   3536  &   $13M_x M_y+13M_x$    &   $13M_x M_y+13M_y$  \\
\cline{2-9}
                & $R_{A}$ &   23.5  &   71  &   238  &   860  &   3256   &    $12M_x M_y+11.5M_x$   &    $12M_x M_y+11.5M_y$ \\
\hline

\textbf{IIR Gabor filter}
\cite{tip/BernardinoS06}  & $R_{M}$ & 34 &   136  &   544  &   2179  & 8704  &     $34M_x M_y$   &     $34M_x M_x$   \\
\cline{2-9}
              & $R_{A}$ & 26 &   104  &   416  &   1664  &   6656  &    $26M_x M_y$    &    $26M_x M_y$   \\
\hline

\textbf{Ours}      & $R_{M}$ & 18 &   54  &   180  &   684  &   2448  &   $9M_x M_y+9M_x$    &   $9M_x M_y+9M_y$  \\
\cline{2-9}
          & $R_{A}$ & 14.5 &  44   &   148  &   536  &   2030  & $7.5M_x M_y+7M_x$    & $7.5M_x M_y+7M_y$  \\
\hline

\hline
\end{tabular}
\end{table*}

Table \ref{tb:SDFT-time} shows the runtime comparison in the
computation of the 2-D localized SDFT. 
It requires computing all 2-D DFT outputs for $u,v=0,...,M-1$, when
$M\times M$ window is used. The 2-D DFT outputs $F_{u,v}$ are
computed by repeatedly applying the methods \cite{tsp/YoungVG02,
tip/BernardinoS06} for $u,v=0....,M-1$. Note that the conjugate
symmetry property, i.e., $F_{u,v}=F^*_{M-u,M-v}$ is used fairly for
all methods when measuring the runtime (see Algorithm
\ref{algo:pseudo_SDFT}). It is clearly shown that our method runs
much faster than the two methods. Interestingly, our runtime gain
against the IIR Gabor filter \cite{tip/BernardinoS06} becomes
higher, when compared to the Gabor filter bank computation in Table
\ref{tb:Gabor-time}. This is mainly because the 1-D horizontal DFT
output $J$ can be reused for $v=0,...M-1$ in the rectangular grid of
Fig. \ref{fig:grid_of_Gabor_SDFT} and it is also shared for both
$M-u$ and $u$ (see Algorithm \ref{algo:pseudo_SDFT}) Namely, the
ratio of shared computations increases in the 2-D localized SDFT.

In Table \ref{tb:SDFT-time}, we also found that the IIR Gabor filter
\cite{tip/BernardinoS06} becomes slower than the recursive Gabor
filter \cite{tsp/YoungVG02} when computing the 2-D localized SDFT,
while the former runs faster than the latter in the Gabor filter
bank computation (compare Table \ref{tb:Gabor-time} and Table
\ref{tb:SDFT-time}). The IIR Gabor filter \cite{tip/BernardinoS06}
decomposes the Gabor kernel into the complex sinusoidal modulation
and the Gaussian kernel, and then perform the Gaussian smoothing
with the modulated 2-D signal. 
Contrarily, the recursive Gabor filter \cite{tsp/YoungVG02} performs
the recursive filtering in a separable manner, and thus we implement
the 2-D localized SDFT using \cite{tsp/YoungVG02} such that it can
reuse 1-D intermediate results, resulting the faster runtime than
\cite{tip/BernardinoS06}. In Table \ref{tb:SDFT-time2}, we also
count the number of multiplications $R_M$ and additions $R_A$, which
are consistent with the runtime results in Table \ref{tb:SDFT-time}.
Here, we also count $R_M$ and $R_A$ when the non-square window of
$M_y\times M_x$ ($M_y \neq M_x$) is used.

\begin{figure}
 \renewcommand{\thesubfigure}{}
 \centering

\hspace{-0.3cm}
\subfigure[(a)]{
\includegraphics[width=0.24\columnwidth]{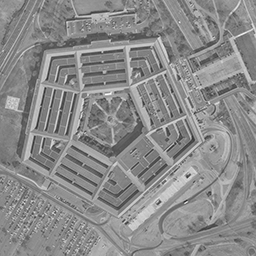}}
\hspace{-0.3cm}
\subfigure[(b)]{
\includegraphics[width=0.24\columnwidth]{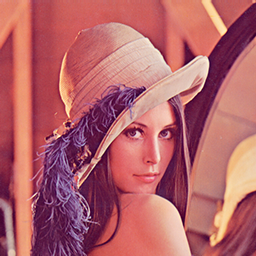}}
\hspace{-0.3cm}
\subfigure[(c)]{
\includegraphics[width=0.24\columnwidth]{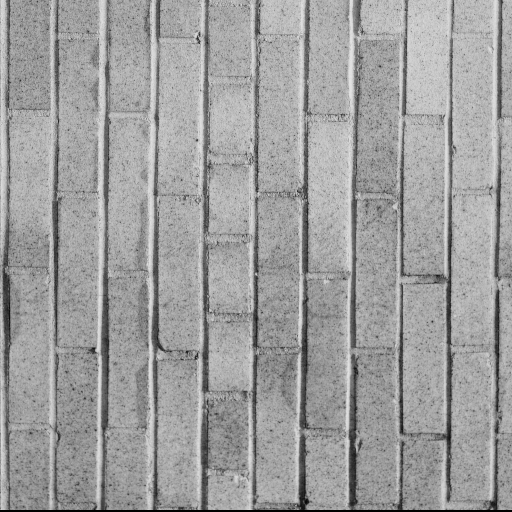}}
\hspace{-0.3cm}
\subfigure[(d)]{
\includegraphics[width=0.24\columnwidth]{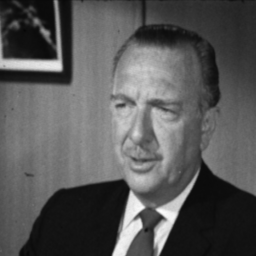}}
\hspace{-0.3cm}

\caption{Some of images used in the experiment (USC-SIPI database
\cite{USC-SIPI}): (a) aerial image, (b) misc image, (c) texture
image, and (d) sequence image.} \label{fig:test_images}
\end{figure}

\begin{figure*}
 \renewcommand{\thesubfigure}{}
 \centering

 \hspace{-1cm}
\subfigure[(a) Aerial]
{\includegraphics[width=0.28\textwidth]{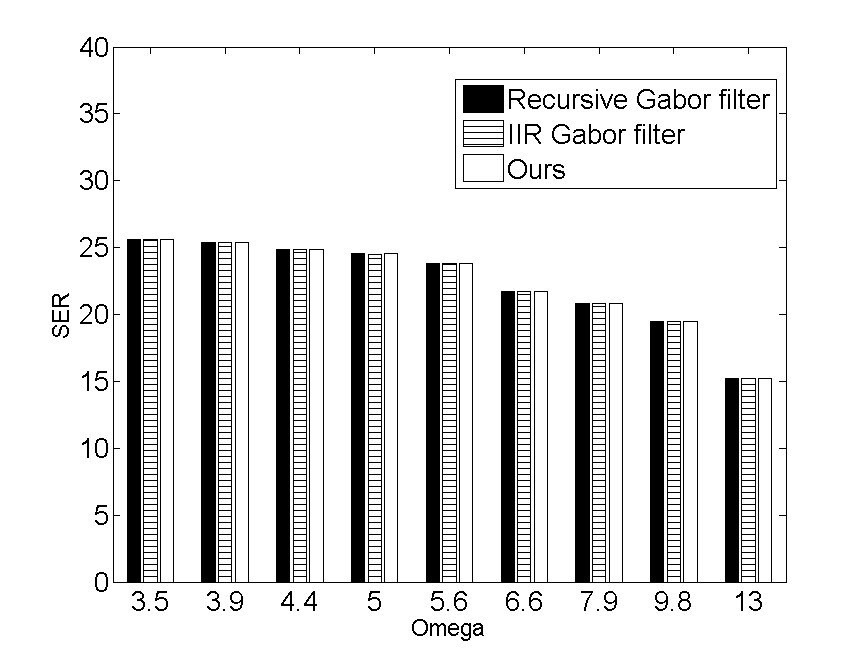}}
 \hspace{-0.7cm}
\subfigure[(b) Misc]
{\includegraphics[width=0.28\textwidth]{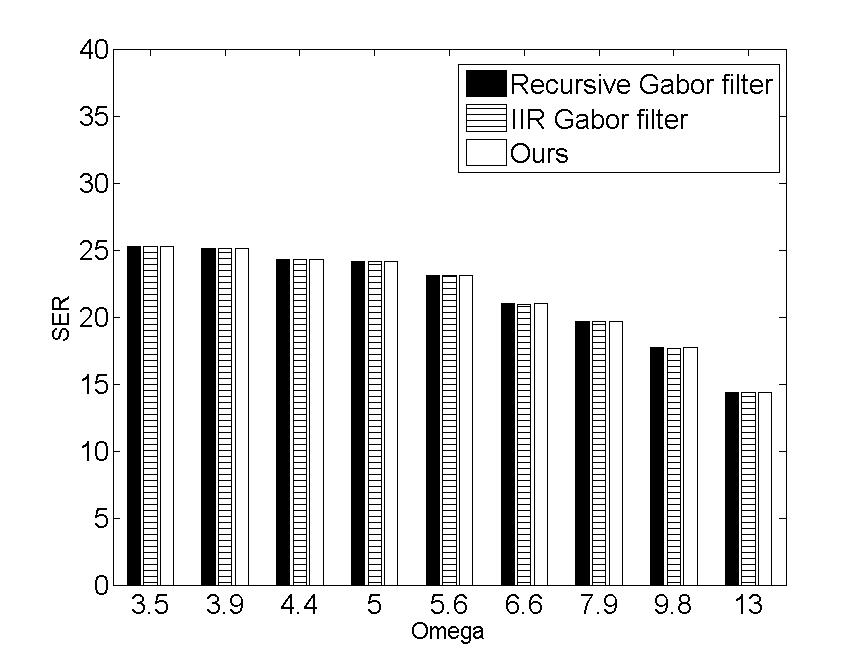}}
 \hspace{-0.7cm}
\subfigure[(c) Sequences]
{\includegraphics[width=0.28\textwidth]{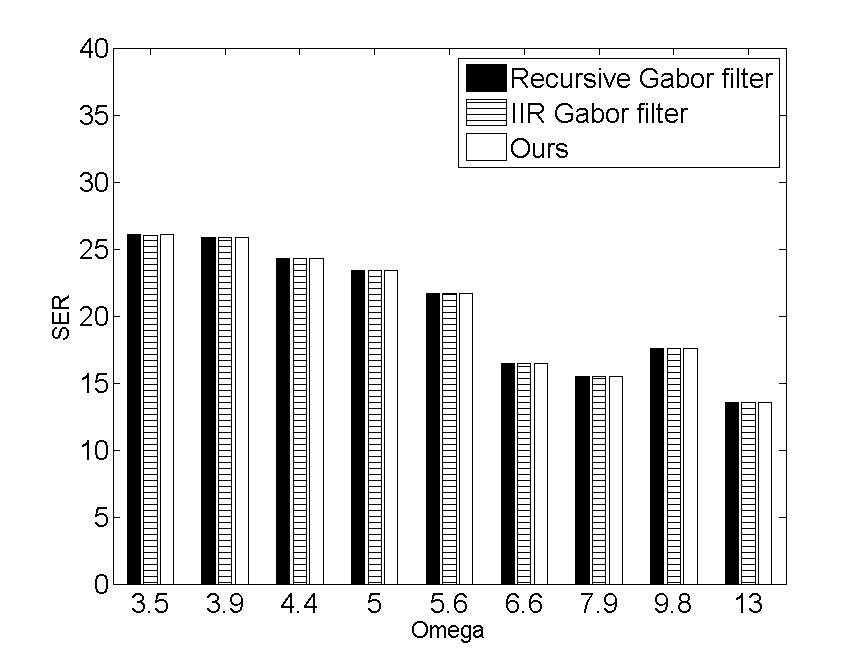}}
 \hspace{-0.7cm}
\subfigure[(d) Textures]
{\includegraphics[width=0.28\textwidth]{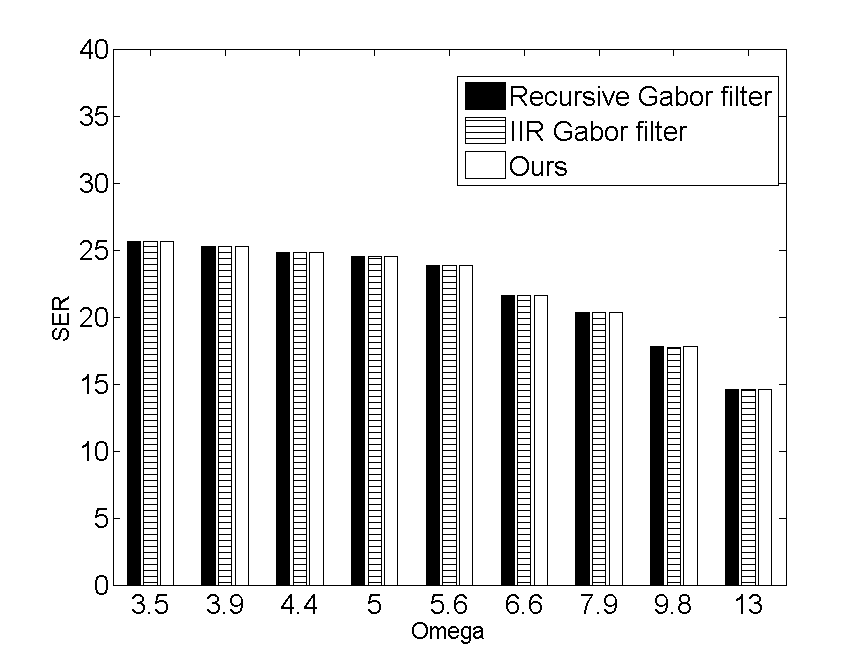}}
 \hspace{-1cm}

\caption{Objective comparison using the imagery parts of 2-D complex
Gabor filtering outputs with the varying frequency $\omega$ when
$\theta=\pi/3$. We compared the average SER values of three methods,
the recursive Gabor filter \cite{tsp/YoungVG02}, IIR Gabor filter
\cite{tip/BernardinoS06}, and our method, for four datasets: (a)
aerial, (b) miscellaneous, (c) sequences, and (d)
textures.}\label{fig:Gabor-SER-imag-freq}
\end{figure*}

\begin{figure*}
 \renewcommand{\thesubfigure}{}
 \centering

 \hspace{-1.1cm}
\subfigure[(a) Aerial]{
\includegraphics[width=0.28\textwidth]{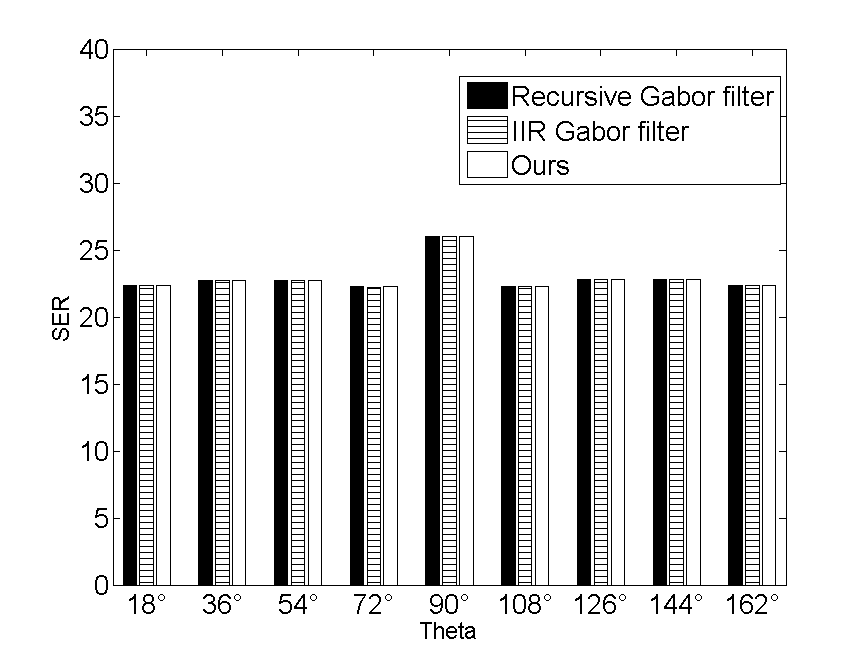}}
 \hspace{-0.72cm}
\subfigure[(b) Misc]{
\includegraphics[width=0.28\textwidth]{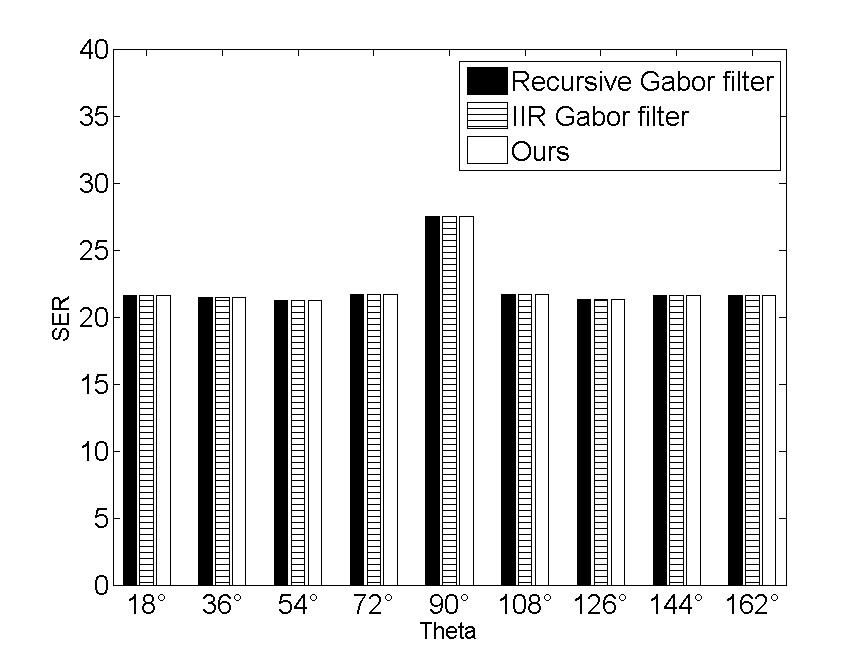}}
 \hspace{-0.72cm}
\subfigure[(c) Sequences]{
\includegraphics[width=0.28\textwidth]{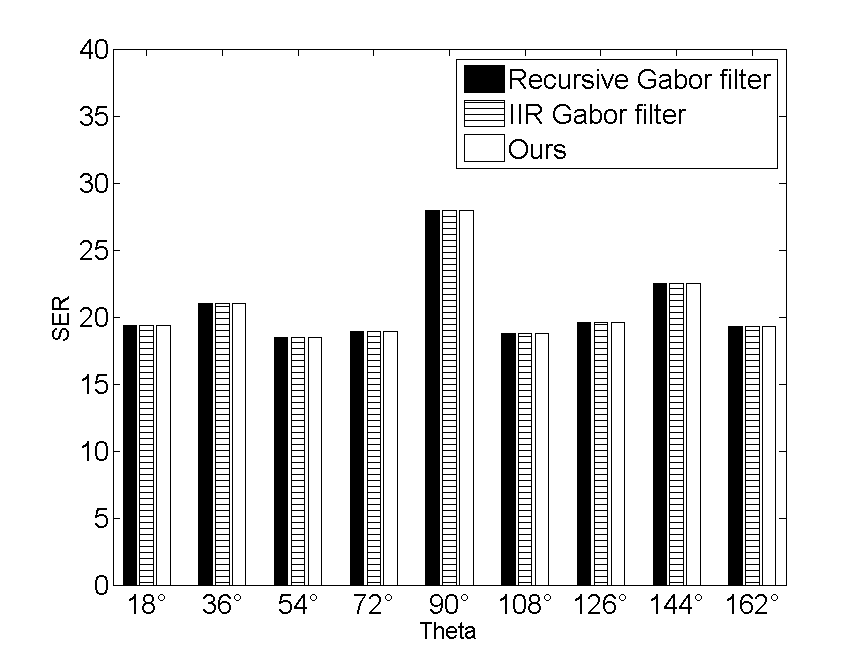}}
 \hspace{-0.72cm}
\subfigure[(d) Textures]{
\includegraphics[width=0.28\textwidth]{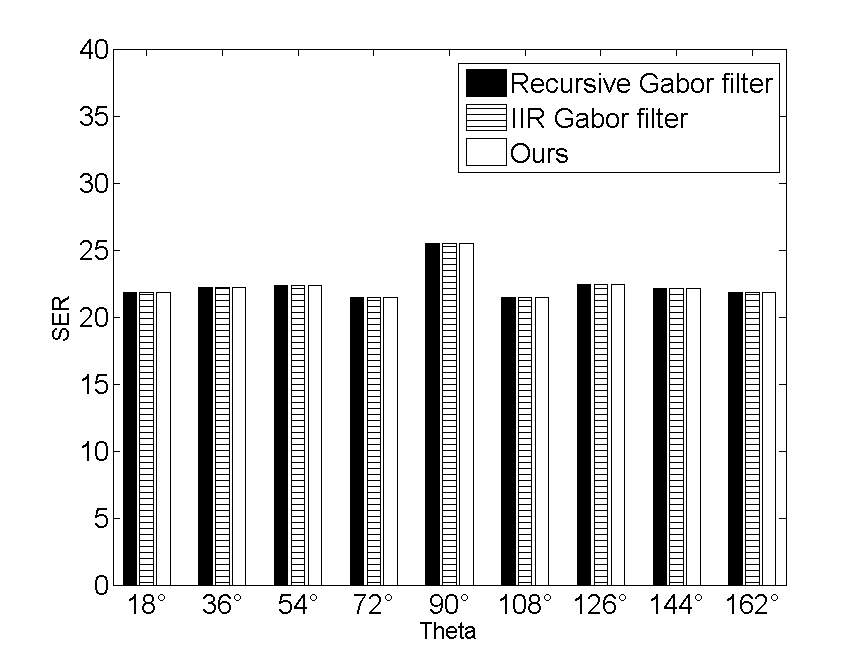}}
 \hspace{-1.1cm}

\caption{Objective comparison using the imagery parts of 2-D Gabor
filtering outputs with the varying orientation $\theta$ when
$\omega=13$. Similar to Fig. \ref{fig:Gabor-SER-imag-freq}, the
average SER values were measured using the recursive Gabor filter
\cite{tsp/YoungVG02}, IIR
Gabor filter \cite{tip/BernardinoS06}, and our method. 
}\label{fig:Gabor-SER-imag-ori}
\end{figure*}

\subsection{Filtering Quality Comparison}
All the fast Gabor filtering methods including ours produce
approximated results, as they counts on the recursive Gaussian
filtering \cite{sigpro/YoungV95,icpr/VlietYV98}. In our method, the
decomposed 1-D signals are convolved using the recursive Gaussian
filtering based on the IIR approximation. The IIR filters run fast
at the cost of the filtering quality loss. It was reported in
\cite{sigpro/YoungV95,icpr/VlietYV98} that the quality loss is
negligible when using the standard deviation within an
appropriate range. 
We compared the filtering quality with two fast Gabor filtering
approaches \cite{tsp/YoungVG02,tip/BernardinoS06}.

We used input images from the USC-SIPI database \cite{USC-SIPI}
which consists of four different classes of images: aerial images,
miscellaneous images, sequence images, and texture images, some of
which are shown in Fig. \ref{fig:test_images}. The filtering quality
was measured for the 2-D complex Gabor filter bank only, as the 2-D
localized SDFT tends to show similar filtering behaviors. We
measured the PSNR by using ground truth results of the
\emph{lossless} FIR Gabor filter in \eqref{eq:original_2d}, and then
computed an objective quality for each of four datasets. The Gabor
filtering outputs are in a complex form, so we measured the
filtering quality for real and imagery parts, respectively. Also,
the filtering outputs do not range from 0 to 255, different from an
image. Thus, instead of the peak signal-to-noise ratio (PSNR) widely
used in an image quality assessment, we computed the signal-to-error
ratio (SER), following \cite{tip/BernardinoS06}:
\begin{equation}
SER[dB] = 10\log _{10} \frac{{\sum\limits_{x,y} {\left( { {\cal R}\{
F(x,y)\} } \right)^2 } }}{{\sum\limits_{x,y} {\left( {{\cal R}\{
F(x,y)\} - {\cal R}\{ F_t (x,y)\} } \right)^2 } }}, \nonumber\\
\end{equation}
where $F$ and $F_t$ are the Gabor filtering results obtained using
the fast method and the lossless FIR filter, respectively. ${\cal
R}(F)$ represents the real part of $F$. The SER can also be measured
with the imagery part ${\cal I}(F)$. We computed the approximation
error for the frequency $\omega \in \{3.5,3.9,...,9.8,13\}$ and the
orientation $\theta \in \{18^\circ,36^\circ,...,162^\circ\}$.

\begin{figure*}
 \renewcommand{\thesubfigure}{}
 \centering

 \hspace{-1cm}
\subfigure[(a) Aerial]
{\includegraphics[width=0.28\textwidth]{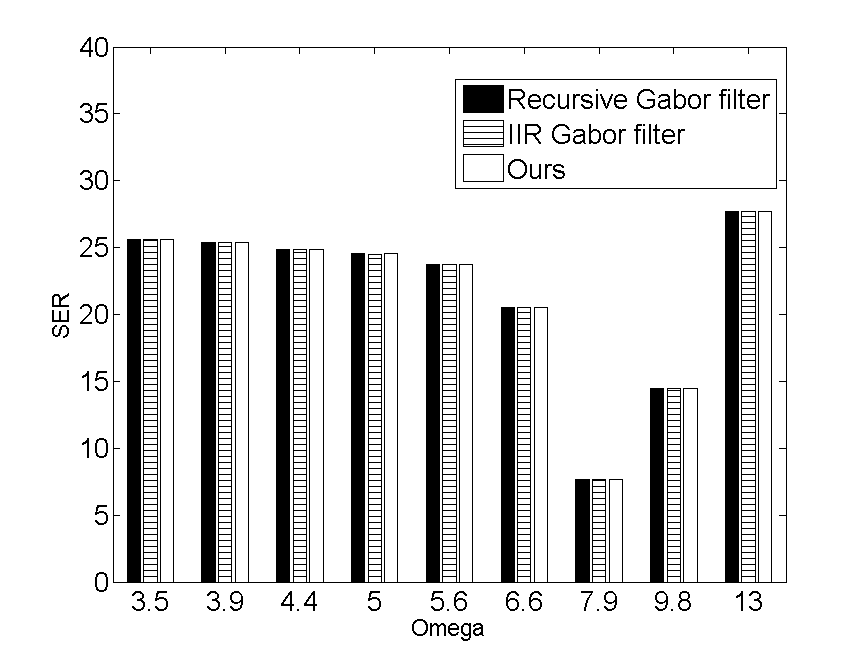}}
 \hspace{-0.7cm}
\subfigure[(b) Misc]
{\includegraphics[width=0.28\textwidth]{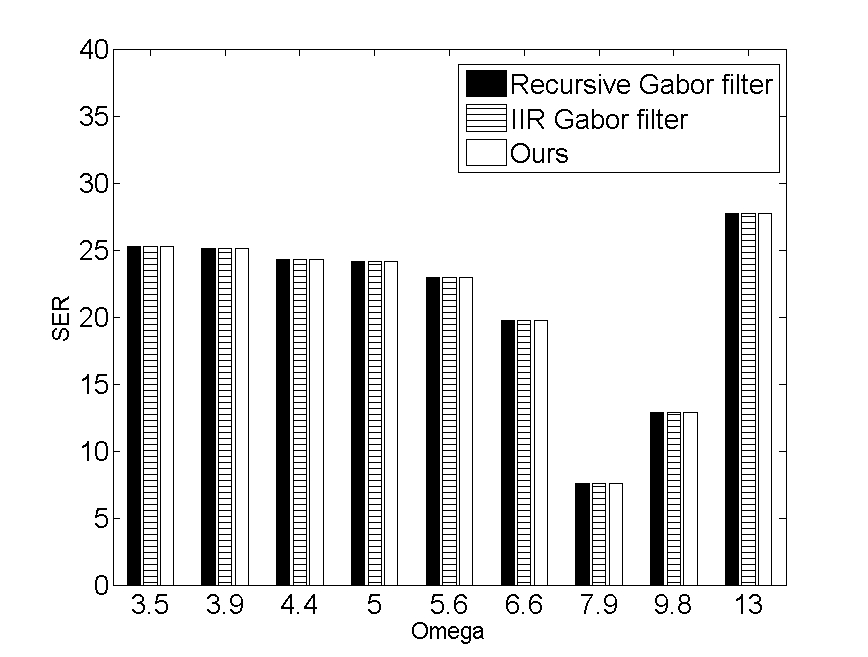}}
 \hspace{-0.7cm}
\subfigure[(c) Sequences]
{\includegraphics[width=0.28\textwidth]{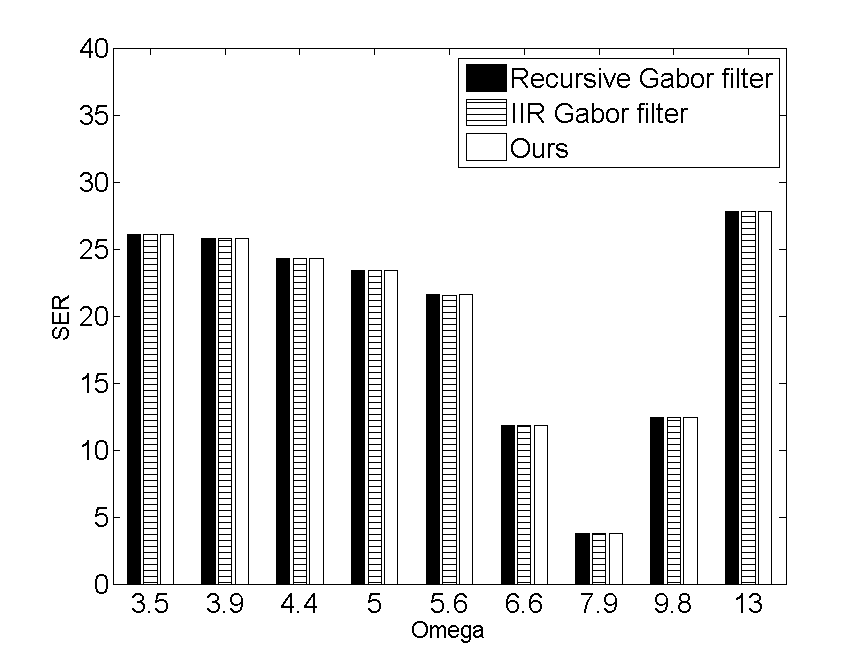}}
 \hspace{-0.7cm}
\subfigure[(d) Textures]
{\includegraphics[width=0.28\textwidth]{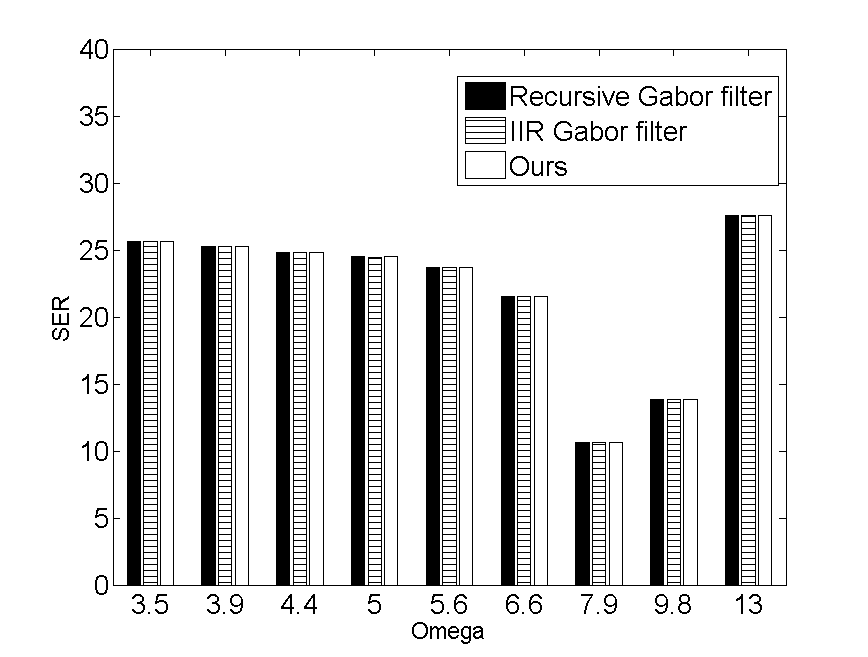}}
 \hspace{-1cm}

\caption{Objective comparison using the real parts of 2-D complex
Gabor filtering outputs with the varying frequency $\omega$ when
$\theta=\pi/3$. The SER values were measured in a manner similar to
Fig. \ref{fig:Gabor-SER-imag-freq}.
}\label{fig:Gabor-SER-real-freq}
\end{figure*}

\begin{figure*}
 \renewcommand{\thesubfigure}{}
 \centering

 \hspace{-1.1cm}
\subfigure[(a) Aerial]{
\includegraphics[width=0.28\textwidth]{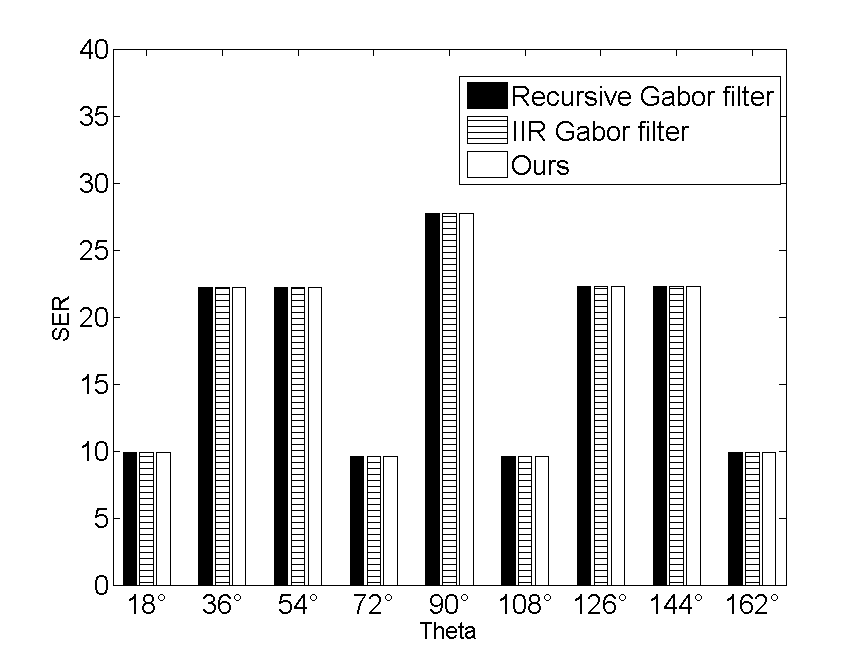}}
 \hspace{-0.72cm}
\subfigure[(b) Misc]{
\includegraphics[width=0.28\textwidth]{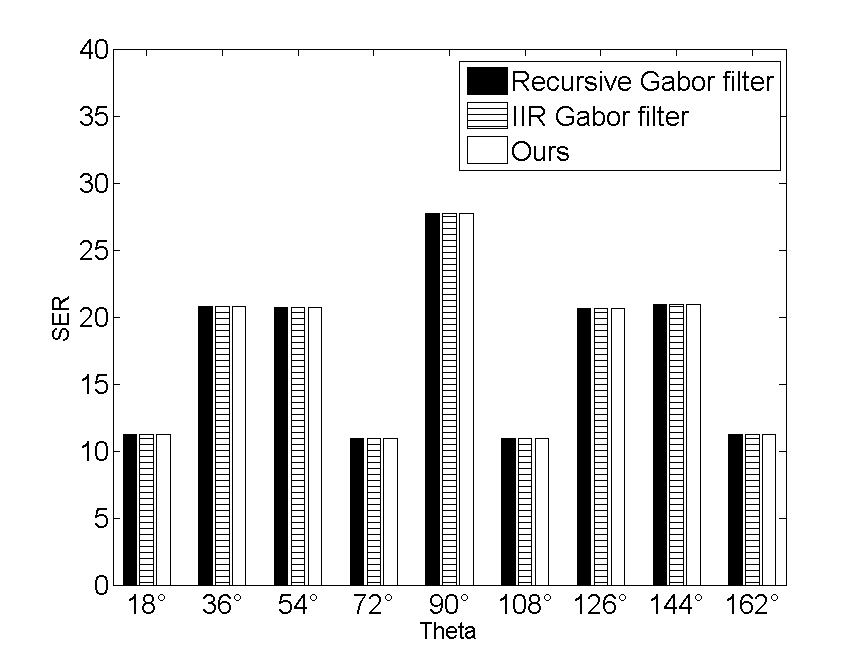}}
 \hspace{-0.72cm}
\subfigure[(c) Sequences]{
\includegraphics[width=0.28\textwidth]{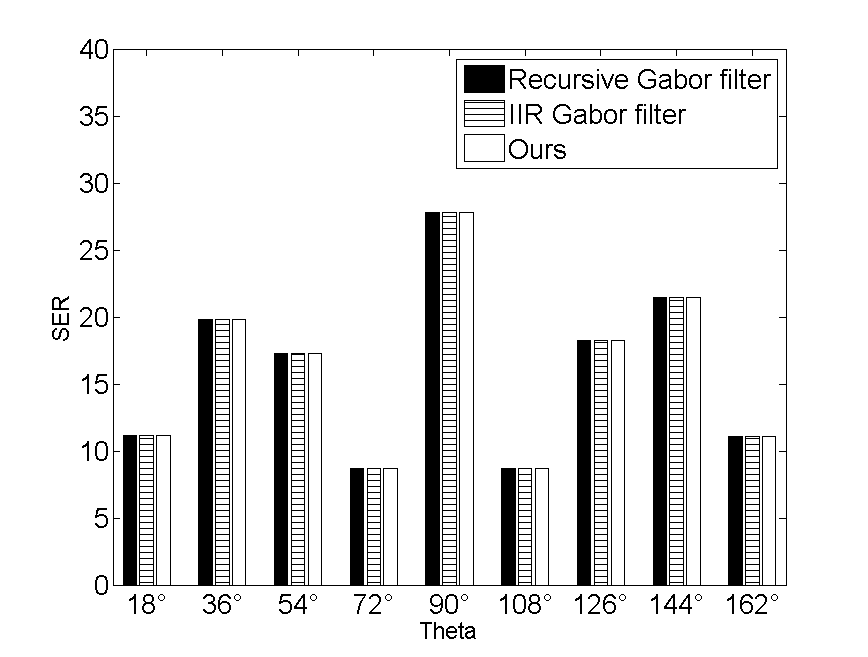}}
 \hspace{-0.72cm}
\subfigure[(d) Textures]{
\includegraphics[width=0.28\textwidth]{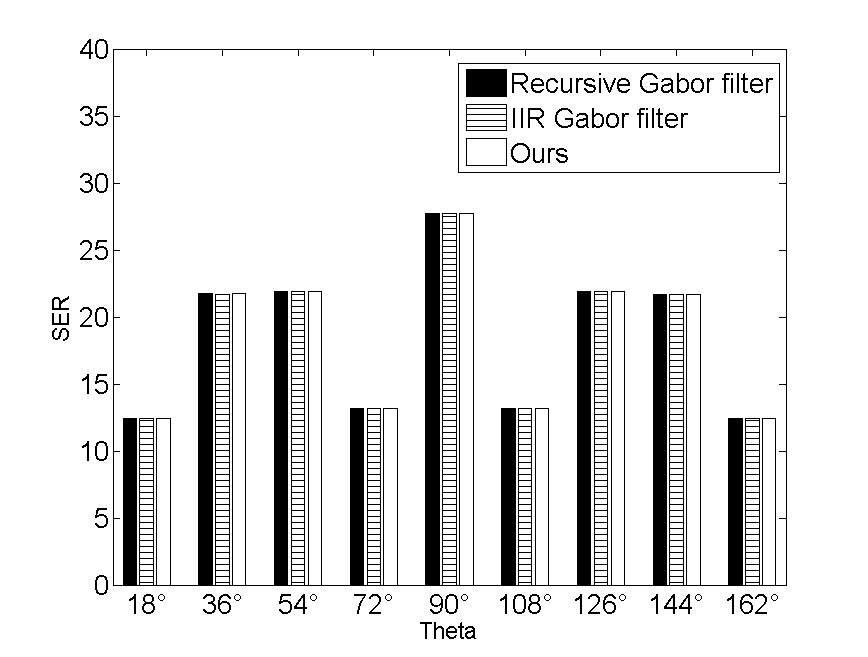}}
 \hspace{-1.1cm}

\caption{Objective comparison using the real parts of 2-D complex
Gabor filtering outputs with the varying orientation $\theta$ when
$\omega=13$. The SER values were measured in a manner similar to
Fig. \ref{fig:Gabor-SER-imag-ori}.
}\label{fig:Gabor-SER-real-ori}
\end{figure*}

\begin{figure}
 \renewcommand{\thesubfigure}{}
 \centering

\subfigure[(a) ]{
\includegraphics[width=0.4\textwidth]{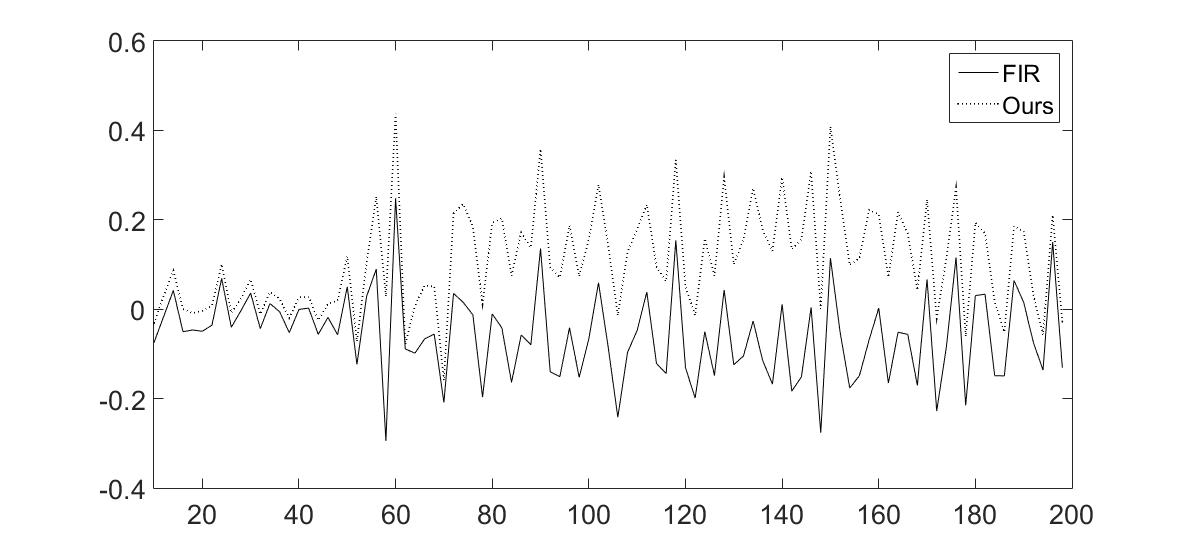}}
\vspace{-0.2cm}

\subfigure[(b) ]{
\includegraphics[width=0.4\textwidth]{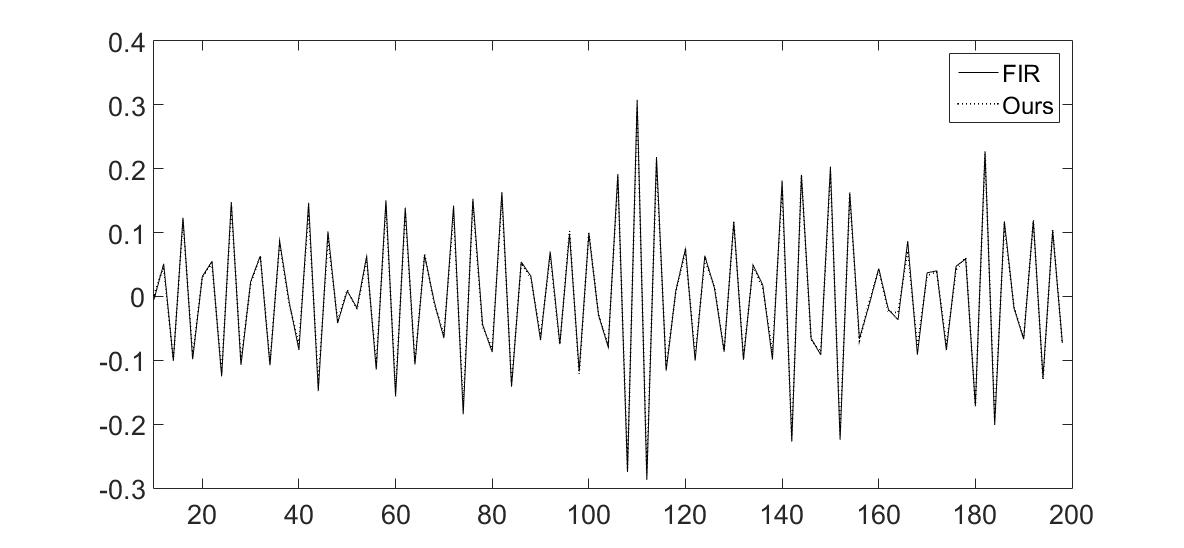}}
\vspace{-0.2cm} \caption{1-D profiles of 2-D complex Gabor filtering
results: (a) the real part at $\omega=7.9$ and $\theta=\pi/3$ when
$SER=10.57$, (b) the real part at $\omega=3.5$ and $\theta=\pi/3$
when $SER=25.61$. }\label{fig:Gabor-profile}
\end{figure}

\begin{figure*}
 \renewcommand{\thesubfigure}{}
  \centering

\subfigure [(a) Face image]
{\includegraphics[width=0.16\textwidth]{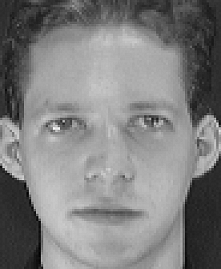}}
\subfigure [(b) $\sigma=3, \theta=\pi/4$]
{\includegraphics[width=0.16\textwidth]{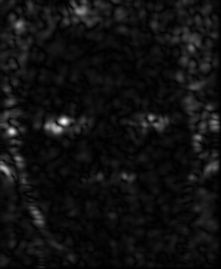}}
\subfigure [(c) $\sigma=3, \theta=3\pi/4$]
{\includegraphics[width=0.16\textwidth]{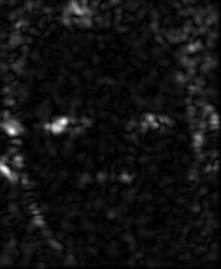}}
\subfigure [(d) $\sigma=2, \theta=0$]
{\includegraphics[width=0.16\textwidth]{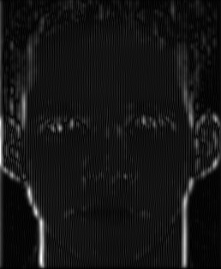}}
\subfigure [(e) $\sigma=2, \theta=\pi/2$]
{\includegraphics[width=0.16\textwidth]{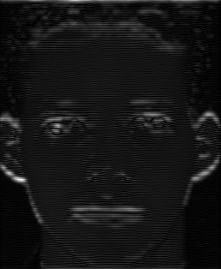}}

\hfill

\subfigure [(f) Texture image]
{\includegraphics[width=0.16\textwidth]{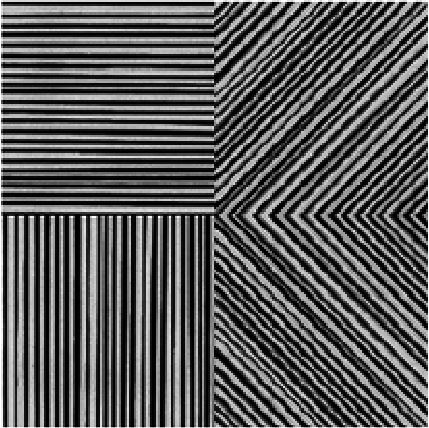}}
\subfigure [(g) $\sigma=2.7, \theta=\pi/4$]
{\includegraphics[width=0.16\textwidth]{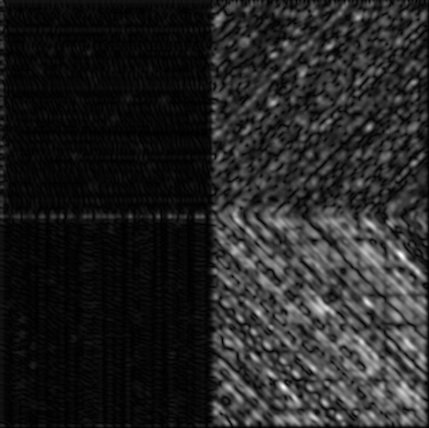}}
\subfigure [(h) $\sigma=2.7, \theta=3\pi/4$]
{\includegraphics[width=0.16\textwidth]{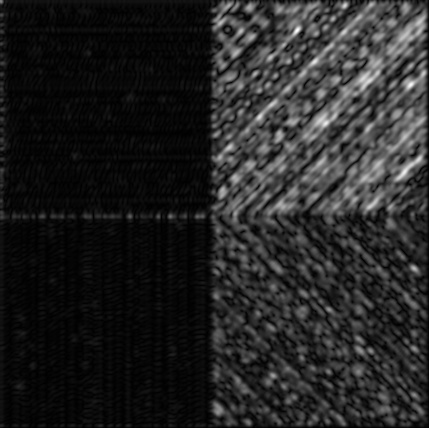}}
\subfigure [(i) $\sigma=4, \theta=0$]
{\includegraphics[width=0.16\textwidth]{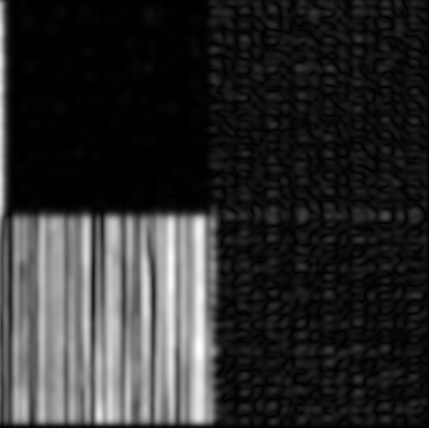}}
\subfigure [(j) $\sigma=4, \theta=\pi/2$]
{\includegraphics[width=0.16\textwidth]{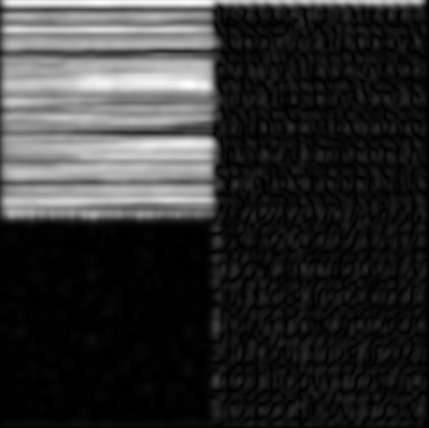}}

\caption{2-D complex Gabor filter bank outputs computed by our
method: The filtering results are in a complex form, so we visualize
them with an absolute magnitude. $\lambda = \sigma/\pi$ depends on
$\sigma$.} \label{fig:Gabor-results}
\end{figure*}

Fig. \ref{fig:Gabor-SER-imag-freq} and \ref{fig:Gabor-SER-imag-ori}
compare the objective Gabor filtering quality by measuring the
average SER values of the imagery parts with respect to the varying
frequency $\omega$ and orientation $\theta$ for four datasets:
aerial,  miscellaneous, sequences, and textures images. The average
SER values are similar to all three methods: the recursive Gabor
filter, IIR Gabor filter, and ours. Four different classes of images
did not show significantly different tendency in terms of the
filtering quality. Fig. \ref{fig:Gabor-SER-real-freq} and
\ref{fig:Gabor-SER-real-ori} shows the SER values measured using the
real parts. Interestingly, the average SER values of the real parts
at some frequencies and orientations become lower. It was explained
in \cite{tip/BernardinoS06} that the difference between DC values of
the lossless FIR and approximated (IIR) filters happens to become
larger at these ranges. In Fig. \ref{fig:Gabor-profile}, we plotted
1-D profiles using the real parts of Gabor filtering results for two
cases with low and high SER values. The horizontal and vertical axes
the pixel location and the real part value of the Gabor filtering,
respectively. In the case with the low SER value, we found that an
overall tendency is somehow preserved with some offsets. Fig.
\ref{fig:Gabor-results} shows the Gabor filtering images obtained
from the proposed method. The absolute magnitude was used for
visualization. Subjective quality of the results are very similar to
that of the original lossless FIR Gabor filtering.

\section{Conclusion}
We have presented a new method for fast computation of the 2-D
complex Gabor filter bank at multiple orientations and frequencies.
By decomposing the Gabor basis kernels and performing the Gabor
filtering in a separable manner, the proposed method achieved a
substantial runtime gain by reducing the computational redundancy
that exists the 2-D complex Gabor filter bank. This method was
further extended into the 2-D localized SDFT that uses the Gaussian
kernel to offer the spatial localization ability as in the Gabor
filter. The computational gain was verified in both analytic and
experimental manners. We also evaluated the filtering quality as the
proposed method counts on the recursive Gaussian filtering based on
IIR approximation. It was shown that the proposed method maintains a
similar level of filtering quality when compared to
state-of-the-arts approaches for fast Gabor filtering, but it runs
much faster. We believe that the proposed method for the fast 2-D
complex Gabor filter bank is crucial to various computer vision
tasks that require a low cost computation. Additionally, the 2-D
localized SDFT is expected to provide more useful information thanks
to the spatial localization property in many tasks based on the
frequency analysis, replacing the conventional 2-D SDFT approaches
using the simple box kernel. We will continue to study the
effectiveness of the 2-D localized SDFT in several computer vision
applications as future work.

\bibliographystyle{IEEEtran}
\bibliography{reference-final}

\end{document}